\documentclass[10pt,journal,compsoc]{IEEEtran}
\usepackage{color}
\usepackage{xcolor}
\usepackage{graphicx}
\usepackage{colortbl}
\usepackage{subfigure}
\usepackage{amsmath,amssymb} 
\usepackage{makecell}
\usepackage{bbding}
\usepackage{tabularx}
\usepackage{booktabs}
\usepackage{subcaption}
\usepackage{textcomp}
\usepackage{wrapfig}
\usepackage{tabu}
\usepackage{multirow}
\usepackage{multicol}
\usepackage{float}
\usepackage{diagbox}
\usepackage{colortbl}
\usepackage{xcolor}
\usepackage{pifont}
\usepackage{algorithm}
\usepackage[noend]{algpseudocode}
\usepackage{titlesec}
\titlespacing*{\paragraph}
  {0pt}{3.25ex plus 1ex minus .2ex}{0em} 

\usepackage[pagebackref=true,breaklinks=true,bookmarks=false]{hyperref}
\usepackage{bm}
\usepackage{url}
\usepackage{tikz}
\usetikzlibrary{spy}

\newcommand{\shortname}{LoRKD }

\begin{document}
\title{LoRKD: Low-Rank Knowledge Decomposition for Medical Foundation Models}

\author{Haolin~Li,
        Yuhang~Zhou,
        Ziheng~Zhao,
        Siyuan~Du,
        Jiangchao~Yao,
        Weidi~Xie,
        Ya~Zhang,
        and Yanfeng~Wang
\IEEEcompsocitemizethanks{
\vspace{-5pt}
\IEEEcompsocthanksitem This work has been submitted to the IEEE for possible publication. Copyright may be transferred without notice, after which this version may no longer be accessible.
\IEEEcompsocthanksitem 
Haolin Li and Siyuan Du are with the School of Computer Science, Fudan University, Shanghai 200437, China, and also with Shanghai
AI Laboratory, Shanghai 200032, China. (E-mail: \{23110240025, 23110240011\}@m.fudan.edu.cn)
\IEEEcompsocthanksitem Yuhang Zhou, Ziheng Zhao, Jiangchao Yao, Weidi Xie, Ya Zhang and Yanfeng Wang are with Shanghai Jiao Tong University, Shanghai 200240, China, and also with Shanghai AI Laboratory, Shanghai 200032, China. (E-mail: \{zhouyuhang, Zhao\_Ziheng, sunarker, weidi, ya\_zhang, wangyanfeng\}@sjtu.edu.cn).
\IEEEcompsocthanksitem Jiangchao Yao and Yanfeng Wang are the corresponding authors.
}}

\markboth{Submitted to IEEE TPAMI}%
{Shell \MakeLowercase{\textit{et al.}}: Bare Advanced Demo of IEEEtran.cls for IEEE Computer Society Journals}

\IEEEtitleabstractindextext{%
\begin{abstract}

The widespread adoption of large-scale pre-training techniques has significantly advanced the development of medical foundation models, enabling them to serve as versatile tools across a broad range of medical tasks.
However, despite their strong generalization capabilities, medical foundation models pre-trained on large-scale datasets tend to suffer from domain gaps between heterogeneous data, leading to suboptimal performance on specific tasks compared to specialist models, as evidenced by previous studies.
In this paper, we explore a new perspective called ``Knowledge Decomposition'' to improve the performance on specific medical tasks, which deconstructs the foundation model into multiple lightweight expert models, each dedicated to a particular anatomical region, with the aim of enhancing specialization and simultaneously reducing resource consumption.
To accomplish the above objective, we propose a novel framework named Low-Rank Knowledge Decomposition (LoRKD), which explicitly separates gradients from different tasks by incorporating low-rank expert modules and efficient knowledge separation convolution.
The low-rank expert modules resolve gradient conflicts between heterogeneous data from different anatomical regions, providing strong specialization at lower costs.
The efficient knowledge separation convolution significantly improves algorithm efficiency by
achieving knowledge separation within a single forward propagation.
Extensive experimental results on segmentation and classification tasks demonstrate that our decomposed models not only achieve state-of-the-art performance but also exhibit superior transferability on downstream tasks, even surpassing the original foundation models in task-specific evaluations.
Moreover, these compact expert models significantly reduce resource consumption, making them more suitable and efficient for practical deployment.
The code is available at \href{https://github.com/MediaBrain-SJTU/LoRKD}{here}.

\end{abstract}

\begin{IEEEkeywords}
Foundation model, Knowledge Decomposition, Low-rank Adaptation, Medical Image Analysis, Universal Pre-training.
\end{IEEEkeywords}}

\maketitle

\IEEEdisplaynontitleabstractindextext

%
\IEEEpeerreviewmaketitle

\section{Introduction}
\label{sec:introduction}

\IEEEPARstart{M}{edical} image analysis powered by deep learning plays a fundamental role in numerous clinical applications, including computer-aided diagnosis, disease progression monitoring, and treatment planning~\cite{nouranian2015learning,yan2018deeplesion, litjens2017survey, zhou2021review, ma2024segment}.
Traditional deep learning models are typically tailored for specific tasks, such as brain tumor segmentation~\cite{havaei2017brain,pereira2016brain,zhao2018deep}.
These models excel only in identifying specific regions of interest (ROI) and exhibit weak adaptability to new tasks, thus can be referred to as ``specialist'' models.
Recently, the research paradigm of medical image analysis has shifted towards universal pretraining~\cite{kirillov2023segment,zhao2023one,ma2024segment,nguyen2023lvm,mei2022radimagenet}, resulting in the development of foundation models, which are pre-trained on large-scale datasets encompassing various anatomical structures and imaging modalities.
These foundation models possess robust transfer and generalization capabilities, allowing them to handle a variety of tasks across different anatomies and modalities.

While foundation models exhibit impressive general feature extraction capabilities, two critical challenges remain in the medical field. 1) 
The significant anatomical differences across various regions of human body, such as the abdomen and brain, incur substantial domain gaps between images from different anatomical structures. 
The cost of pretraining on such heterogeneous data usually comes with sacrificing the performance of individual regions. 
Specifically, recent studies in medical field~\cite{glocker2023risk, huang2024segment,wu2023can} have shown that the performance of foundation models remains inferior to that of specialist methods, implying that current medical foundation models may not be able to well guarantee both generality and specialization simultaneously. 2) Foundation models, characterized by their extensive parameters and high computational demands, are impractical for deployment in diverse resource-constrained medical environments~\cite{bommasani2021opportunities,xu2024survey,sun2023dime,touvron2023llama}. 
For example, as highlighted in~\cite{moor2023foundation,zhou2024reprogramming,zhang2024generalist}, medical foundation models require high-performance hardware that is often difficult for hospitals to acquire, particularly for hospitals in underdeveloped areas.

\begin{figure}[t]
    \centering
    \includegraphics[width=\linewidth]{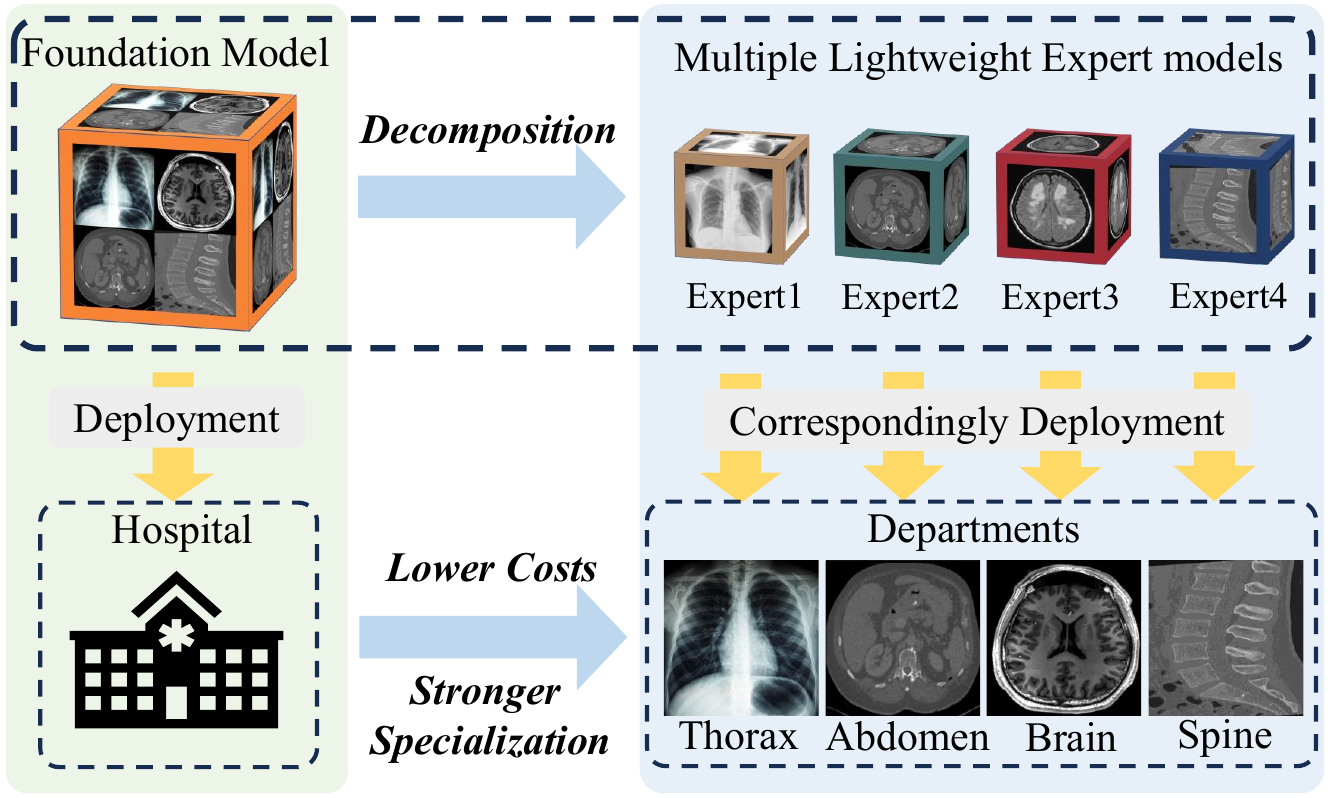}
    \caption{Knowledge decomposition is employed to break down the foundation model into multiple lightweight expert models, each tailored to a specific domain. The goal of this paradigm is to improve the specialization of deployment models within a specific domain, while simultaneously reducing deployment costs.}
    \label{fig:background}
    \vspace{-0.4cm}
\end{figure}

To address the aforementioned issues, we propose a new perspective called knowledge decomposition, which aims to offer potential solutions for the practical application of cumbersome medical foundation models.
The objective of knowledge decomposition is to decompose the foundation model into multiple lightweight expert models, where each expert model concentrates exclusively on a specific region, following the department taxonomy of a hospital (as shown in Figure~\ref{fig:background}). 
1) In contrast to specialist models that are designed to handle single specific task of a particular region (such as segmenting lung tumors in thoracic imaging), the decomposed expert models we decompose are capable of managing \emph{all tasks} within their respective regions. 
For instance, an expert model dedicated to the thorax can perform segmentation tasks across various organs and conditions within the thoracic region, such as those involving the lungs, heart, and other thoracic structures.
2) Compared to foundation models that tackle all tasks across all regions, decomposed expert models effectively mitigate conflicts arising from heterogeneous data, leading to \emph{enhanced specialization and reduced deployment costs.}
To the best of our knowledge, there has been no research conducted in the medical field on how to decompose a foundation model into multiple expert models. 
The most related study in the field of natural images, KF~\cite{yang2022factorizing}, has made preliminary explorations into this problem. 
KF factorizes the pre-trained model into a common knowledge network (CKN) and several task-specific networks (TSNs) by manipulating the mutual information between models. 
After decomposition, the CKN can be combined with each TSN to form task-specific expert models. 
However, the indefinite primary-secondary structure design requires trivial training and cannot effectively decouple knowledge from different regions solely by means of the loss function. Regarding the lightweight aspect, the introduction of TSNs also results in significant resource overhead, making the approach inefficient for practical applications.

In this work, we propose Low-Rank Knowledge Decomposition (LoRKD), a method for decomposing the medical foundation model into lightweight, task-specific experts.
Our \shortname consists of two main components: low-rank expert modules and efficient knowledge separation convolution.
Concretely, the low-rank expert modules comprise two main modules: a primary common-shared backbone and secondary task-specific low-rank expert modules \emph{attached} to the backbone.
The common-shared backbone, which houses the majority of the model parameters, is utilized to learn generic knowledge shared across all tasks. 
The region-specific low-rank expert modules employ low-rank adapters (LoRA)~\cite{hu2021lora} to assimilate domain-specific knowledge, explicitly segregating gradients from different regions into their corresponding modules. 
This architecture efficiently controls parameter growth while resolving conflicts in heterogeneous data.
However, such a design also comes along with a critical technical challenge: when a mini-batch contains data from multiple tasks, the forward operation should be performed multiple times, which greatly increases training time.
To address this issue, we introduce efficient knowledge separation convolution to achieve knowledge separation at the convolutional level.
This approach enables gradients to be separated into their corresponding expert modules in a single forward propagation, while simultaneously accumulating them in the shared backbone.
Furthermore, considering the varying difficulty of tasks in different regions, we present two variants,  LoRKD and LoRKD*.
The former sets the same rank for each region's low-rank expert module.
The latter implements an automated, imbalanced design for the ranks of different expert modules. Specifically, for regions with more challenging tasks, the rank of their expert modules is set higher to enhance the representation ability; conversely, for regions with relatively simpler tasks, the rank of their expert modules is set lower to reduce costs.

During inference, the low-rank expert modules can be integrated into the backbone, further reducing inference latency and computational overhead.
For scenarios necessitating targeted analysis of a particular region, only the relevant low-rank module needs to be fused with the common backbone to create an expert model. 
For instance, in the thoracic surgery department, only the thorax expert module is required. 
This integrated model, compared to the original foundation model, boasts fewer parameters and superior performance, thereby achieving cost reduction and
performance improvement simultaneously.
In a nutshell, our contributions are summarized as follows:
\begin{itemize}
\item \textbf{Knowledge Decomposition.}
Given the significant data heterogeneity in medical area, we introduce knowledge decomposition to broaden the application of medical foundation models, which decomposes foundation models into multiple lightweight experts to reduce costs and enhance specialization. 
\item \textbf{Novel Framework.}
We introduce a novel method LoRKD, which comprises two components: low-rank expert modules and the efficient knowledge separation convolution. 
LoRKD injects task-specific knowledge into the corresponding expert modules via efficient explicit gradient separation.
\item \textbf{Superior Performance.}
Extensive experiments on both segmentation and classification tasks demonstrate the superiority of our method.
\shortname can decompose the foundation models into lighter yet stronger expert models, leading to superior specialization and transferability to downstream tasks.
Comprehensive analysis further verifies the potential and applicability of knowledge decomposition.

\end{itemize}

\section{Related Work}
\label{sec:related_work}

\subsection{Medical Foundation Models}

Medical foundation models powered by universal pre-training have emerged as a crucial advancement in medical image analysis, driving significant progress across various tasks.
Pretrained on large-scale diverse datasets, medical foundation models exhibit remarkable performance and generality.
These models can be broadly categorized into those designed for segmentation tasks and those for diagnosis tasks.
To advance segmentation foundation models, researchers have undertaken preliminary explorations.
Several methods have concentrated on fine-tuning SAM~\cite{kirillov2023segment} on medical data~\cite{ma2024segment, cheng2023sam, wu2023medical,huang2024segment}, 3DSAM-adapter~\cite{gong20233dsam} and SAM-Med3D~\cite{Wang2023SAMMed3D} introduce novel methods to adapt SAM from 2D natural images to 3D volumetric images, fully leveraging spatial information.
Other works have explored alternative pre-trained models~\cite{ye2023uniseg, liu2023clip}.
SAT~\cite{zhao2023one} employed knowledge-enhanced representation learning to pre-train universal segmentation model with text prompts, while UniverSeg~\cite{butoi2023universeg} exploited a CrossBlock mechanism to learn precise segmentation maps without additional training.


Similarly, foundation models for disease diagnosis exhibit strong generality and transferability to downstream classification tasks~\cite{xie2018pre,zhang2016automatic,shin2016deep}. 
Recent advancements in classification foundation models have explored various pre-training methods.
Some studies attempt to develop a medical version of ImageNet to facilitate the pre-training of medical image classification models~\cite{mei2022radimagenet,yang2023medmnist,dai2024unichest}, thereby enhancing the transferability of pre-trained models to downstream tasks.
Other works concentrate on designing self-supervised pre-training methods tailored for medical images~\cite{azizi2021big,tang2022self,chaves2022evaluation,krishnan2022self,zhou2023unified}.
Furthermore, some studies leverage text information, including medical records and terminology descriptions, to develop advanced multimodal pre-training algorithms~\cite{li2020comparison,moon2022multi,huang2021gloria,chen2022multi,taleb2022contig,liang2021contrastive}.


Despite these advances, all these foundation models face challenges such as gradient conflicts and high computational costs, particularly when trained on large-scale medical image datasets spanning various body regions and anatomies~\cite{yuan2022decentralized,yu2020gradient,senushkin2023independent}. 

\begin{figure}[t]
    \centering
    \includegraphics[width=\linewidth]{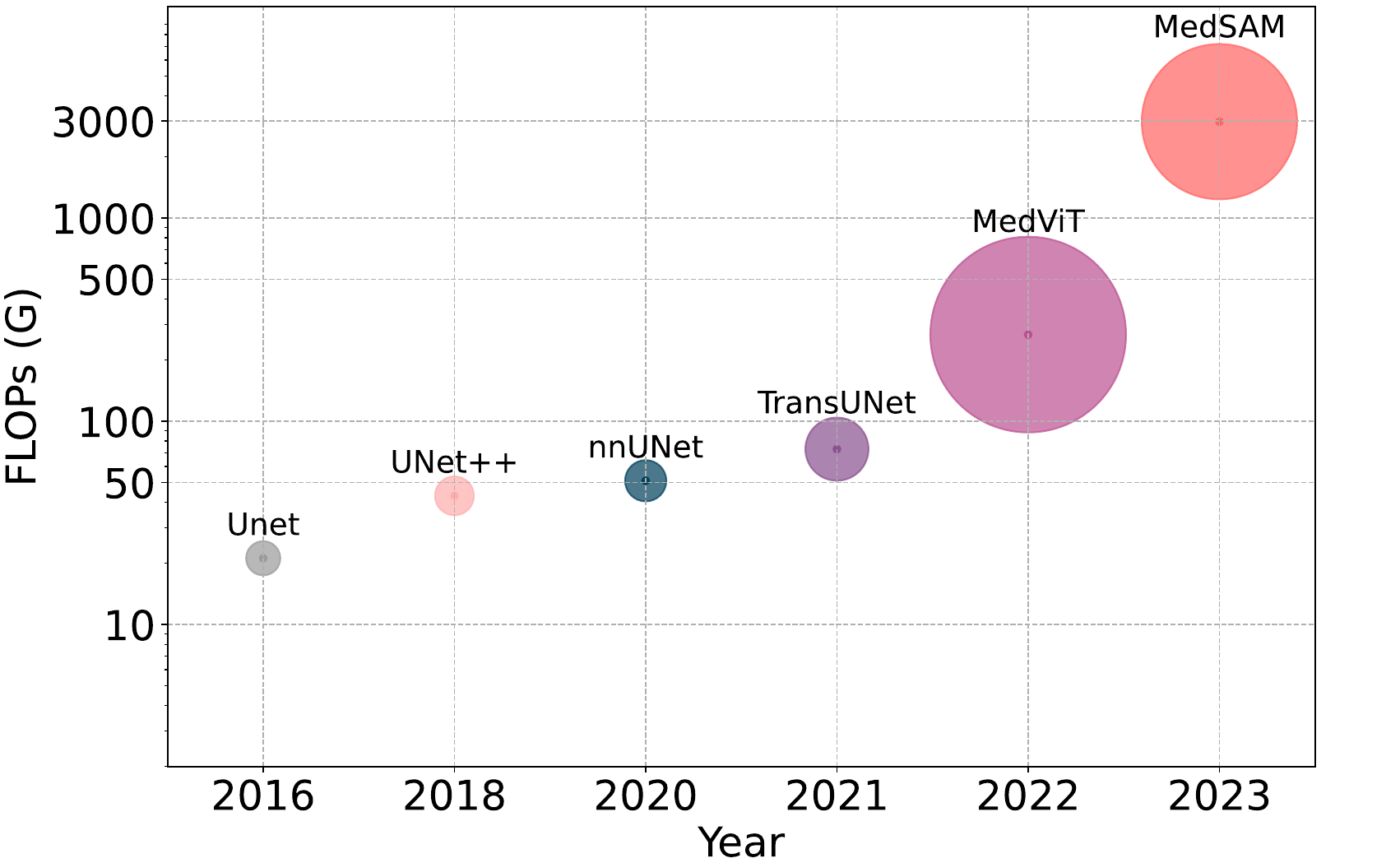}
    \caption{The resource consumption of foundation models is growing at an exponential rate.
    The size of the circle represents the model's parameters.
    }
    \label{fig:scale}
    \vspace{-0.3cm}
\end{figure}

\subsection{Knowledge Decomposition}
Different from the previous disentangled representation learning that is usually done through variational auto-encoder~\cite{burgess2018understanding, higgins2016beta, kim2018disentangling} or adversarial learning~\cite{tran2017disentangled, chen2016infogan, liu2018multi, mathieu2016disentangling}, the goal of knowledge decomposition is to break down the pre-trained foundation model into multiple region-specific experts.
Recently, in the field of natural images, KF~\cite{yang2022factorizing} conducted early exploration of knowledge decomposition by promoting modularization of knowledge through optimizing mutual information loss~\cite{hjelm2018learning, lowe2019greedy,oord2018representation}. 
It factorizes a pre-trained model into a common knowledge network and several task-specific networks. 
In this work, we conduct the first exploration of knowledge decomposition in the medical field and propose a novel approach that not only better controls the number of parameters but also attains a more advanced level of performance and transferability.

\subsection{Low-Rank Adaptation}
Low-Rank adaptation (LoRA) is a parameter-efficient fine-tuning method for large language models~\cite{hu2021lora}.
During fine-tuning, LoRA utilizes low-rank matrices to approximate the changes in pre-trained weights.
The low-rank matrices can be re-parameterized into the pre-trained weights to avoid inference latency.
Due to its impressive performance and efficiency, many LoRA variants have been proposed~\cite{kopiczko2023vera, zhang2023adaptive, chen2023longlora,hayou2024lora+}.
QLoRA~\cite{dettmers2024qlora} combined LoRA with 4-bit NormalFloat quantization to further reduce computational costs.
DoRA~\cite{liu2024dora} decomposed the weight change into magnitude and direction components and utilized LoRA to fine-tune the direction component.
Galore ~\cite{zhao2024galore} implemented a gradient low-rank projection method to reduce optimizer memory usage, allowing full-parameter training under limited resources.
These LoRA variants are designed solely for parameter-efficient fine-tuning, while our \shortname employs low-rank structures as knowledge carriers for specific tasks to alleviate conflicts between heterogeneous data and simultaneously maintain minimal growth in model parameters.
\begin{figure}[t]
    \centering
    \includegraphics[width=0.77\linewidth]{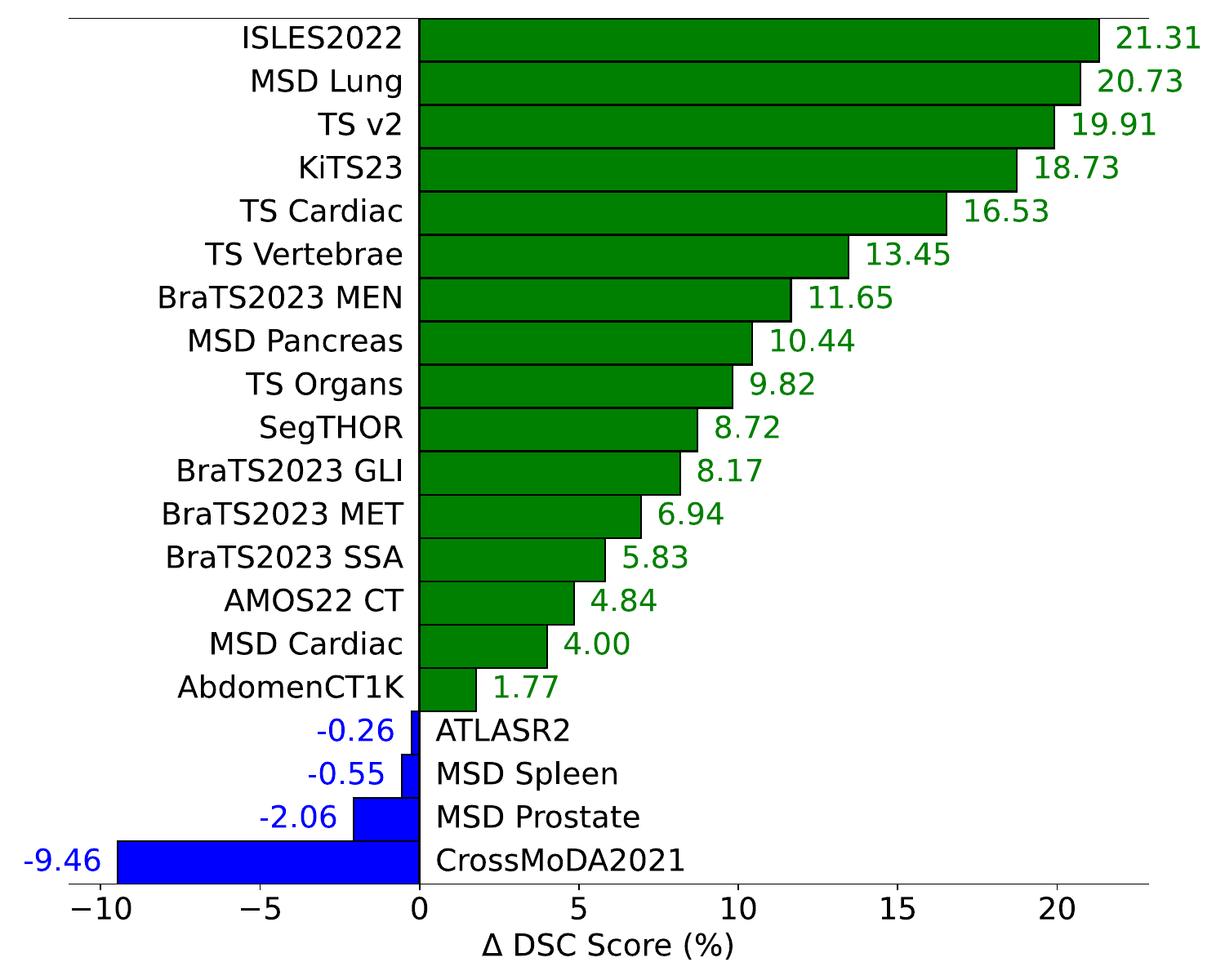}
    \caption{Performance comparison between the foundation model and specialist model. $\triangle$ DSC is the DSC value of nnUNet minus the DSC value of MedSAM.
    }
    \label{fig:motivation}
    \vspace{-0.3cm}
\end{figure}



\section{Methodology}\label{sec:method}

In this section, we first present the problem formulation and motivation in \S\ref{preliminary} and \S\ref{motivation} respectively. 
Then, we introduce the details of our \shortname in three parts: \S\ref{molora} describes the low-rank expert modules; \S\ref{eksconv} presents the efficient knowledge separation convolution; the training objective of \shortname is shown in \S\ref{loss}; we provide the decomposition procedure and analysis algorithm complexity in \S\ref{procedure}.

\subsection{Preliminary}
\label{preliminary}

\begin{figure*}[t]
    \centering
    \includegraphics[width=\linewidth]{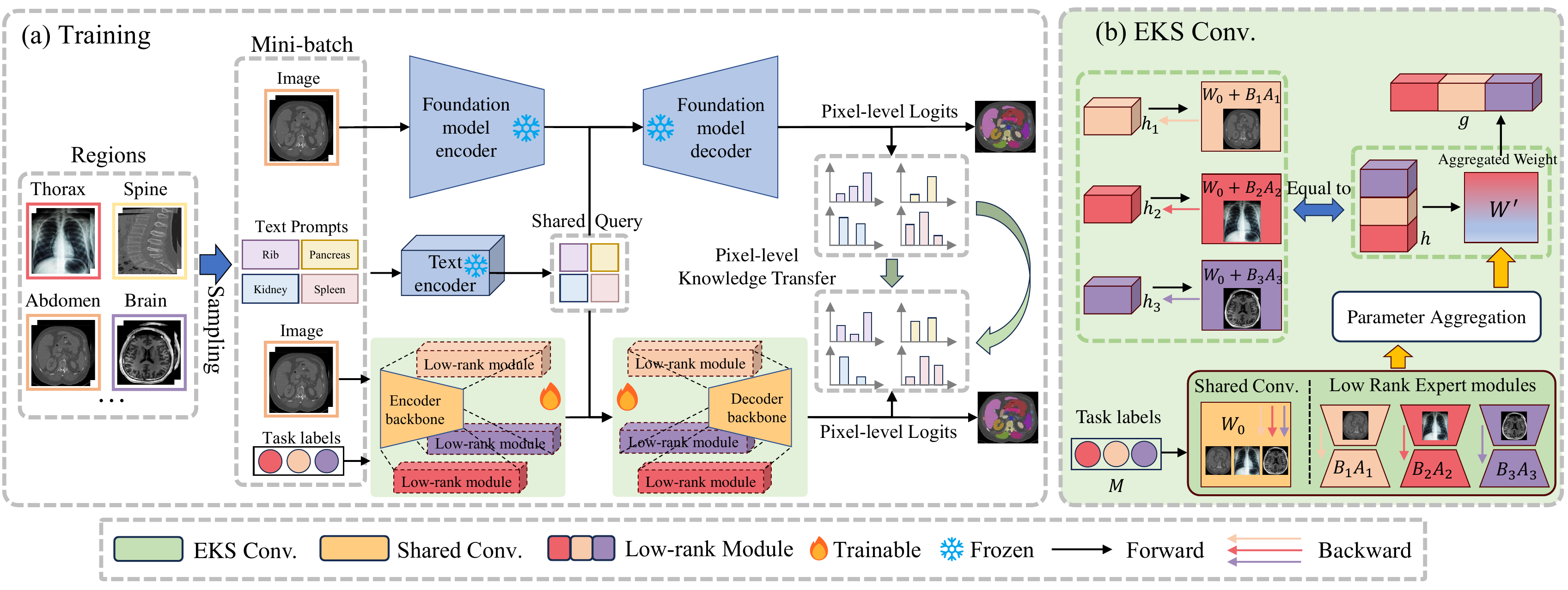}
    \caption{The illustration of LoRKD for medical foundation models on segmentation. The low-rank expert modules control the number of parameters and efficient knowledge separation convolution (EKS Conv) achieves computationally efficient gradient separation. Decomposed models can replace medical foundation model in specific domains and can switch task knowledge conveniently between departments. The case for classification tasks holds by turning the decoders as classifiers.
    }
    \vspace{-0.4cm}
    \label{fig:method}
\end{figure*}

Considering that medical images are predominantly volumetric and 3D images inherently contain richer contextual information compared to 2D images, our method is presented from a 3D perspective for simplicity. 
Note that, our method can be naturally compatible with 2D cases by simply degenerating the input dimension.
Assuming we have a universal pretraining dataset $D=\{(x_1, y_1),...,(x_n, y_n)\}$, where n is the number of data, $x_i\in\mathbb{R}^{C\times H\times W \times D}$ represents the input volumetric image, and $y_i$ is the prediction target.
$C, H, W, D$ is the channel, height, width, and depth of the feature maps, respectively.
For the segmentation task, $y_i\in\mathbb{R}^{K\times H\times W \times D}$ is the binary segmentation masks of the
anatomical targets and $K$ stands for the number of segmentation targets.
For the classification task, $y_i\in\{0,1,...K-1\}$ is the class label of $x_i$ and $K$ is the number of classes.


Given a foundation model $F$ pre-trained on heterogeneous datasets covering multiple anatomical regions, our goal is to decompose $F$ into several lightweight models $F_{1},..., F_{T}$, where each lightweight model is an expert model corresponding to a specific anatomical region. 
Specifically, our decomposed model $F_{d}$ consists of a common-shared backbone $F_s$ and $T$ low-rank expert modules $E_{1},..., E_{T}$, with each expert module specializing in a particular region, such as the brain or abdomen. 
An expert model $F_{i}$ can be obtained by compositing the low-rank expert module $E_{i}$ with the shared backbone, namely, $F_{i}=F_s \circ E_i$.

\subsection{Motivation}
\label{motivation}




The increasing size of foundation models has led to significant challenges regarding computational resources and efficiency.
Figure~\ref{fig:scale} illustrates the growth trend in the number of parameters and computational requirements (measured in FLOPs) for well-known medical models.
As it is shown, while these models excel at general feature extraction, their massive parameter counts demand substantial computational power, making them impractical for many real-world scenarios. 
Additionally, despite their generality, foundation models often underperform compared to specialist models on specific medical tasks.
As shown in Figure~\ref{fig:motivation}, we evaluated the performance of a state-of-the-art foundation model MedSAM~\cite{ma2024segment} against a state-of-the-art specialist model nnUNet~\cite{isensee2021nnu} on 20 distinct datasets. 
Our results showed that the specialist model outperformed the foundation model in most cases, achieving superior results on 16 out of 20 datasets. 
This further demonstrates the lack of specialization in foundation models for medical tasks.

To reduce costs and enhance specialization, we propose our LoRKD, which tackles these two issues from the perspective of knowledge decomposition.
LoRKD consists of two main components: the low-rank expert modules and the efficient knowledge separation convolution.
We explicitly separate the gradients from different anatomical regions into corresponding low-rank expert modules.
Our intuition is that the expert modules can then learn task-specific knowledge while the shared backbone can acquire general knowledge, thus resolving gradient conflicts between heterogeneous data.
To handle the computational challenge posed by multiple expert modules, we introduce efficient knowledge separation convolution, which enables gradient separation to be accomplished in a single forward pass, significantly reducing computational overhead. 
Besides, during the inference for specific regions, the composition of expert modules and shared backbone makes the parameter size in a tolerable scale compared to medical foundation models. The overall framework of our \shortname is illustrated in Figure~\ref{fig:method}.

\subsection{Low-Rank Expert Modules}
\label{molora}
Considering the limited computational resources and the scalability required for numerous tasks, expert modules carrying region-specific knowledge need to strike a balance between the number of parameters and feature representation capability.
LoRA~\cite{hu2021lora}, a widely used fine-tuning method in large language models, has been demonstrated to be parameter-efficient~\cite{zhang2023lora,valipour2022dylora}.
Inspired by this, we propose to use a similar low-rank structure as the carriers for knowledge decomposition, named low-rank expert modules.

Given a shared convolution $\mathbf{W_0}\in \mathbb{R}^{C^{\text{out}}\times C^{\text{in}}\times k\times k\times k}$ in $F_s$, where $C^{\text{out}}, C^{\text{in}}, k$ represent the number of output channels, the number of input channels, and the kernel size respectively. 
We configure two low-rank factors $\mathbf{B_t}\in \mathbb{R}^{C^{\text{out}}k\times rk}$ and $\mathbf{A_t} \in \mathbb{R}^{rk\times C^{\text{in}}k^{2}}$ for $t$-th expert, where $r$ represents the rank. 
As a result, for the features belonging to the $t$-th task, original convolution operation $g_t = \mathbf{W_0}h_t$ can be transformed into:
\begin{align}
\label{equ:m1}
g_t = (\mathbf{W_0}+\mathbf{B_t}\mathbf{A_t}) h_t, 
\end{align}
where, for brevity, we omit the reshape operation, and $h_t$, $g_t$ represent the input features and output features respectively.
It is worth noting that, different from previous scenarios where $\mathbf{W_0}$ remains frozen in LoRA, in our knowledge decomposition scenario, $\mathbf{W_0}$, as a carrier of common knowledge, requires to be updated along with the low-rank factors $\mathbf{A_t}$ and $\mathbf{B_t}$.

\begin{figure}[t]
    \centering
    \subfigure[SAT-DS]{
        \centering
        \includegraphics[width=0.23\textwidth]{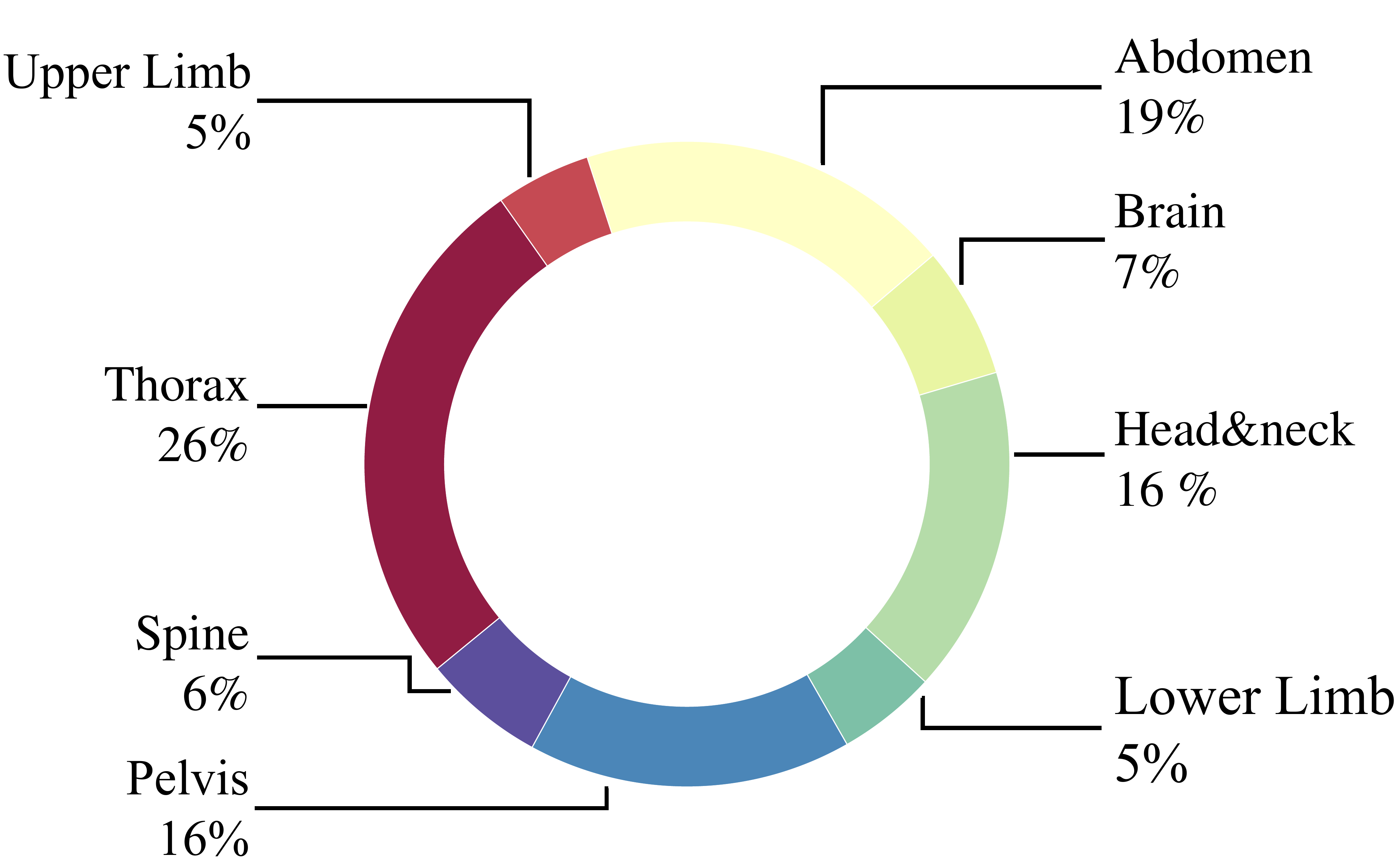}
        
        \label{fig:im-sat}
    }
    \subfigure[Radimagenet]{
        \centering
        \includegraphics[width=0.23\textwidth]{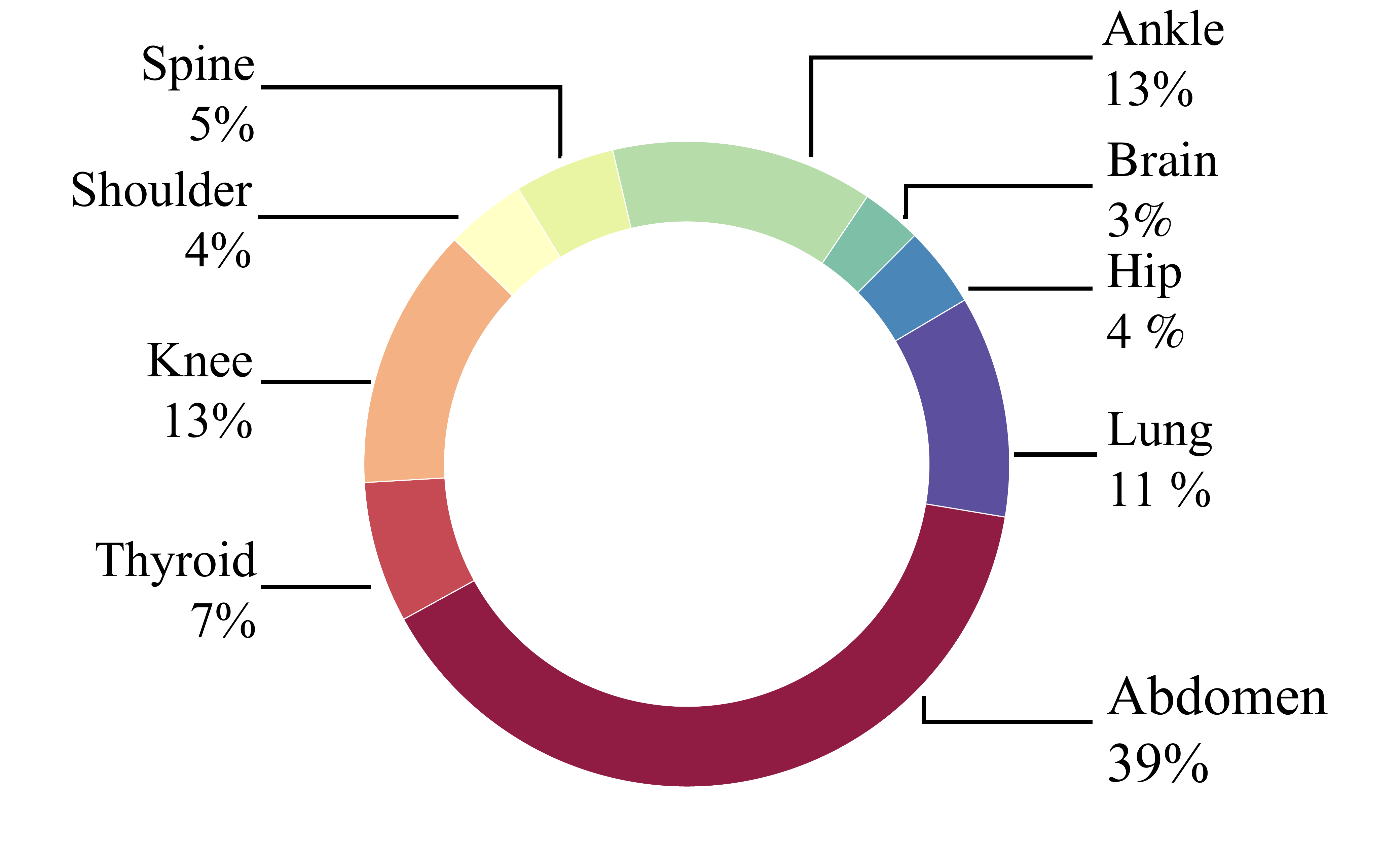}
        \label{fig:im-rad}
    }
    \caption{Data distribution in two large-scale medical datasets.
    }
    \vspace{-0.4cm}
    \label{fig:imbalance}
\end{figure}

\subsubsection{Task disparity requires imbalanced rank design}
\label{imbalance-rank}

The intrinsic differences between various anatomical regions present substantial challenges in medical image analysis using neural networks.
These variations arise from several factors, including distinct anatomical features, tissue densities, potential pathologies, and the specific imaging modalities employed~\cite{moor2023foundation}. 
For instance, the differences between medical imaging of the brain and the thorax are significant. 
Brain imaging is predominantly performed using MRI, which provides detailed images of soft tissues and is crucial for identifying neurological conditions~\cite{balafar2010review,moeskops2016automatic}. 
In contrast, imaging of the thorax often employs CT or X-ray modalities, which are better suited for visualizing dense structures and detecting conditions related to the lungs~\cite{dai2024unichest,ccalli2021deep}.
Additionally, the data employed for universal pre-training is highly imbalanced, with most images coming from a few regions, as illustrated in Figure~\ref{fig:imbalance}. 
This imbalance exacerbates the difficulty of tasks associated with underrepresented regions, as the neural network's training is skewed towards more frequently imaged areas. Therefore, the difficulty level of tasks across different anatomical regions is markedly disparate, necessitating tailored approaches to adaptively address this unique challenge.

Specifically, for regions with a large loss reduction during the warmup phase, the corresponding low-rank expert modules are assigned larger rank values. 
A large loss reduction indicates significant optimization space, necessitating a larger rank for sufficient representation capabilities. 
Conversely, regions with a small loss reduction during the warmup phase have limited common knowledge and require more task-specific knowledge to compensate. 
The low-rank expert module needs to be sufficiently differentiated from the backbone to allow the task-specific knowledge to develop a distinct representation separate from the common knowledge. 
The smaller the rank of the low-rank expert module, the more differentiated it is from the backbone; as the rank increases and reaches that of the backbone, they become equivalent.
Therefore, the low-rank expert modules associated with these regions are assigned smaller rank values to ensure their differentiation from the backbone.

To adapt to the varying difficulties across different regions, we devise LoRKD*, a variant of our method. 
LoRKD* adaptively adjusts the ranks of the low-rank modules through an automated mechanism.
Specifically, we quantify the changes in the loss function of data from different regions during the warmup phase and adjust the ranks of the corresponding low-rank expert modules accordingly.
Assume that the loss reduction of each region during the warmup phase is $\Delta \mathcal{L}_1,...,\Delta \mathcal{L}_T$, and the base rank is $r$. 
The rank of each low-rank module can then be calculated as:
\begin{align}
\label{equ:warmup}
r_{i}=\lfloor r\cdot(\frac{\Delta\mathcal{L}_{i}}{\Delta\mathcal{L}_{avg}})^{2} \rfloor_e, 
\end{align}
where $\Delta\mathcal{L}_{avg}$ is the average of $\Delta \mathcal{L}_1,...,\Delta \mathcal{L}_T$ and $\lfloor x \rfloor_e$ denotes rounding $x$ to the nearest even number.

\subsection{Efficient Knowledge Separation Convolution}
\label{eksconv}


\par{\noindent \bf Task-Specific Gradient Separation Bottleneck.}
To achieve knowledge decomposition, we propose explicit gradient separation as our solution. 
This approach ensures that each expert module computes gradients exclusively for its designated task, thus acquiring task-specific knowledge.
Concurrently, the shared backbone aggregates gradients from all tasks, thereby acquiring generic knowledge shared across all tasks.
However, when a mini-batch of data contains $T$ tasks, \emph{the convolution operation must be performed $T$ times $g_t = (\mathbf{W_0}+\mathbf{B_t}\mathbf{A_t}) h_t$, where $t\in \{1,...,T\}$. }
The $T$ times forward propagation significantly increases the training time, especially when decomposing a large number of tasks. 
To address this issue, we propose the Efficient Knowledge Separation Convolution (EKS Convolution).

In order to elucidate our improvements in convolution, we first review the standard convolution operation. 
For each convolution, the input features can be represented as $h \in \mathbb{R}^{B \times C^{\text{in}} \times H \times W \times D}$, where $B, H, W, D$ represent the sample number of mini-batch size, the height, width, and depth of the feature maps, respectively. 
If the kernel size of the convolution is $k$ and the stride is 1, each output feature unit $o_{ijl} \in \mathbb{R}^{B \times C^{\text{out}}}$ in output features $g \in \mathbb{R}^{B \times C^{\text{out}} \times H \times W \times D}$ can be expressed as
\begin{align}
o_{ijl}=\sum_{m=0}^{k-1}\sum_{n=0}^{k-1}\sum_{o=0}^{k-1} h_{(i+m)(j+n)(l+o)} \cdot \omega_{mno}, \nonumber
\end{align}
where $i \in \{ 1, ..., H \}$ , $j \in \{ 1, ..., W \}$ , $l \in \{ 1, ..., D \}$ , and $h_{(i+m)(j+n)(l+o)} \in \mathbb{R}^{B\times C^{\text{in}}}$ represents the units of the input feature map $h$, while $\omega_{mno} \in \mathbb{R}^{C^{\text{in}} \times C^{\text{out}}}$ represents the convolution weights.

For each EKS Convolution, in addition to the input feature map $h$, the task label $\mathbf{M} \in \mathbb{R}^{B\times  T}$, which is a one-hot vector corresponding to the mini-batch, is also inputted as a reference for subsequent parameter aggregation.
The output features are then computed as follows:
\begin{equation}
\setlength{\abovedisplayskip}{3pt}
\setlength{\belowdisplayskip}{3pt}
    \begin{aligned}
    g&=g_1\cup\cdots\cup g_t \cup\cdots\cup g_T\\
    g_t&=(\mathbf{W_0}+\mathbf{B_t}\mathbf{A_t})h_t =(\mathbf{W_0}+\mathbf{B_t}\mathbf{A_t})\mathbf{M_t}h, 
\end{aligned}
\label{eq:forward}
\end{equation}
where $\cup$ denotes the concatenation operation, $h_t $ represents the set of $B^t$ features in $h$ that correspond to the $t$-th task, and  $\mathbf{M_t}$ is an index matrix that indicates which features in $h$ belong to the $t$-th task. 
To avoid redundant convolutional operations, we propose parameter aggregation, wherein the parameters for the current iteration are aggregated into $\mathbf{W^{\prime}}$ according to $\mathbf{M}$. 
This ensures that the number of forward propagation is always equal to 1, and the operation $g=\mathbf{W^{\prime}}h$ is equivalent to the Eqn.~\eqref{eq:forward}. Specifically, the operation of the Eqn.~\eqref{eq:forward} can be transformed as follows:
\begin{equation}
\setlength{\abovedisplayskip}{3pt}
\setlength{\belowdisplayskip}{3pt}
\begin{aligned}
\label{eks}
    g & =(\mathbf{W_0}+\mathbf{B_1}\mathbf{A_1})h_1\cup\cdots\cup (\mathbf{W_0}+\mathbf{B_T}\mathbf{A_T})h_T \\
    &=(\mathbf{W_0}+\sum\nolimits_{i=1}^{T} (\mathbf{\widetilde{BA}}\odot \mathbf{M})_i)h =\mathbf{W^{\prime}} h,
\end{aligned}
\end{equation}
where $\mathbf{\widetilde{BA}} \in \mathbb{R}^{1\times T\times C^{\text{out}}\times C^{\text{in}} \times k \times k \times k}$ contains the weights of all low-rank expert modules, which can be obtained by
\begin{equation}
    \mathbf{\widetilde{BA}} = \mathbf{B_1}\mathbf{A_1} \cup ...\cup \mathbf{B_t}\mathbf{A_t}\cup...\cup \mathbf{B_T}\mathbf{A_T}.\nonumber
\end{equation}
$\odot$ denotes the Hadamard product, and $\mathbf{\widetilde{BA}}\odot \mathbf{M} \in \mathbb{R}^{B\times T\times C^{\text{out}}\times C^{\text{in}} \times k \times k \times k}$ represents the configuration of low-rank expert module for each input feature and $i$ corresponds to the second dimension of $(\mathbf{\widetilde{BA}}\odot \mathbf{M})$. The weight of shared convolution $\mathbf{W_0}$ is applied to all tasks.
In this way, we obtain the aggregated weight $\mathbf{W^{\prime}} \in \mathbb{R}^{B\times C^{\text{out}}\times C^{\text{in}} \times k \times k \times k}$ that is equivalent to Eqn.~\eqref{eq:forward} but requires only single forward.

Another challenge associated with it is that $\mathbf{W^{\prime}}$ has six dimensions, unlike traditional 3D convolutions which typically have five dimensions. 
To ensure compatibility with existing deep learning libraries, we adopted the concept of group convolution (GConv)~\cite{krizhevsky2012imagenet}.
Specifically, we set the group number to $B$ and $\gamma\in \{1,...,B\}$. Then, we reshape $h$ to $h \in \mathbb{R}^{1\times BC^{\text{in}}\times H\times W\times D}$ and reshape $\mathbf{W^{\prime}}$ to $\mathbf{W^{\prime}}\in \mathbb{R}^{BC^{\text{out}}\times C^{\text{in}}\times k\times k\times k}$. Consequently, each output feature unit $o_{ijl}$ in $g$ can be computed by 
\begin{equation}
\setlength{\abovedisplayskip}{3pt}
\setlength{\belowdisplayskip}{3pt}
\begin{aligned}
    o_{ijl}&=o_{ijl}^1\cup\cdots\cup o_{ijl}^\gamma\cup\cdots\cup o_{ijl}^B \\
o_{ijl}^\gamma&=\sum_{m=0}^{k-1}\sum_{n=0}^{k-1}\sum_{o=0}^{l-1}h_{(i+m)(j+n)(l+o)}^\gamma  \cdot \omega_{mno}^\gamma, 
\end{aligned}
\label{gconv}
\end{equation}
where $h_{(i+m)(j+n)(l+o)}^\gamma$ and $\omega_{mno}^\gamma$ represent the reshaped versions.
Eqn.~\eqref{gconv} is a standard form of group convolution, which can be easily implemented in existing deep learning libraries such as PyTorch~\cite{paszke2019pytorch} and TensorFlow~\cite{abadi2016tensorflow}. With the above transformations, EKS Convolution improves upon the traditional convolution operation by enabling gradient separation to be achieved in a single forward pass, regardless of the number of tasks. Besides, it eliminates the computational overhead of duplicating input for each convolution, thereby significantly improving training efficiency.

\subsection{Training Objective}
\label{loss}
For objective, we design distinct loss functions specific to medical foundation models for segmentation and classification tasks.
In general, the loss function of \shortname comprises two main parts: $\mathcal{L}_{task}$ and $\mathcal{L}_{transfer}$.
$\mathcal{L}_{task}$ provides supervision from the label information of the corresponding task, while $\mathcal{L}_{transfer}$ transfers knowledge from the foundation model to decomposed models, which can be expressed as:
\begin{align}
\label{loss-seg}
\mathcal{L}_{\mathrm{total}}=\mathcal{L}_{task}+\beta \mathcal{L}_{\mathrm{transfer}},
\end{align}
where $\beta$ is a trade-off hyperparameter.

\subsubsection{Training Objective for Segmentation}
For medical foundation models towards segmentation, following~\cite{isensee2021nnu}, we employ dice loss and binary cross-entropy loss as $\mathcal{L}_{task}$.
Specifically, given a sample with $K$ classes and $C$ voxels, 
the decomposed model prediction and ground-truth are denoted as $p_{c,k}$ and $s_{c,k}$ respectively, and then we can formulate $\mathcal{L}_{task}$  as follows:
\begin{equation}
\begin{aligned}
\mathcal{L}_{task}&=\mathcal{L}_{bce}+\mathcal{L}_{dice} \\
&=-\frac{1}{K}\sum_{k=1}^{K}\frac{1}{C}\sum_{c=1}^{C}p_{c,k}\cdot \log s_{c,k} + \\
&(1-\frac{2\sum_{k=1}^{K}\sum_{c=1}^{C}p_{c,k}\cdot s_{c,k}}{\sum_{k=1}^{K}\sum_{c=1}^{C}p_{c,k}^{2}+\sum_{k=1}^{K}\sum_{c=1}^{C}s_{c,k}^{2}})
\end{aligned}
\end{equation}
In order to transfer the knowledge from the medical foundation model into the lightweight decomposed models, we directly distill fine-grained knowledge at the predicted mask level.
Let $p_{c,k}^{b}$ denote the prediction of the foundation model, and then $\mathcal{L}_{\mathrm{transfer}}$ can be computed as:
\begin{equation}
\begin{aligned}
\mathcal{L}_{\mathrm{transfer}}&=\mathcal{L}_{\mathrm{KL}}(p_{c,k}^{b}, p_{c,k}) \\
&=\sum_{k=1}^{K}\sum_{c=1}^{C}p_{c,k}^{b} \log \frac{p_{c,k}^{b}}{p_{c,k}},
\end{aligned}
\end{equation}
where $\mathcal{L}_{\mathrm{KL}}$ represents the Kullback-Leibler divergence.

\subsubsection{Training Objective for Classification}
For medical foundation models towards the classification task, given a mini-batch of training data $\{(x_i,y_i, y_i^t)\}_{i=1}^B$, $x_i$ represents the $i$-th input image in the current mini-batch, $y_i$ represents the class label across all tasks and $y_i^t$ represents the class label within its corresponding task $t$.
We denote the feature extracted from the foundation model as $f^b_i = F (x_i; \theta_{F})$, and the features extracted from the lightweight decompostion model as $f^d_i = F(x_i; \theta_{F_s};\theta_{E_t})$.
Then, the $\mathcal{L}_{\mathrm{transfer}}$ for sample $x_i$ can be written as $\mathcal{L}_{\mathrm{KL}}(f_{i}^b,f_{i}^d)$.

Moreover, we can also leverage class label information $\{y_i^t\}$ to enhance task-level supervision. Specifically, during training, we integrate $T$ classification heads $\{h_1,...,h_T\}$ into the lightweight decompostion model. These classification heads can individually predict $\{Y_1,...,Y_T\}$ classes where $Y_t$ represents the number of classes for the $t$-th task, $Y$ is the total number of all classes and $\sum_{i=1}^TY_i=Y$.
The logits extracted from the decomposition model can be denoted as $g^d_i = h_t (f^d_i)$ and the prediction  can be calculated by:
\begin{equation}
p_{i,j}^d = \frac{\exp(g^d_{ij}/\tau)}{\Sigma_{j=1}^{Y_t}\exp(g^d_{ij}/\tau)},\nonumber
\end{equation}
where $g^d_{ij}$ represents the $j$-th logit in $g^d_i$ and $\tau$ is the temperature. $\mathcal{L}_{\mathrm{CE}}(y_i^t,p^d_{i})$ represents the task-level supervision loss of $x_i$. Then, the total loss of a mini-batch can be written as:
\begin{equation}
\begin{aligned}
\label{loss-cls}
\mathcal{L}_{\mathrm{total}}&=\frac{1}{B}\sum_{t=1}^{T}\sum_{i=1}^{B^t}\left [ \mathcal{L}_{\mathrm{CE}}(y_i^t,p^d_{i})+\beta\mathcal{L}_{\mathrm{KL}}(f_{i}^b,f_{i}^d)\right ].
\end{aligned}
\end{equation}

\subsection{Algorithm and Complexity}
\label{procedure}

\begin{algorithm}[!t]
	\small
	\begin{algorithmic}[1]
		\renewcommand{\algorithmicrequire}{\textbf{Input:}}
		\renewcommand{\algorithmicensure}{\textbf{Output:}}
\Require Dataset $D=\{(x_1, y_1),...,(x_n, y_n)\}$ and the foundation model $F$
		\Ensure Predicted target $Y^d$
        \State  Initialize the network parameters and hyper-parameters such as $\beta, r$
		\If {LoRKD-imbalance}
        \State Warmup training, calculate the rank for each region according to Equ.(\ref{equ:warmup})
        \ElsIf {LoRKD-balance}
        \State Warmup training, set each low-rank expert module with the base rank $r$
        \EndIf
        \For {each step}
        \State Foundation model forward $y^b_i=F(x_{i})$, where $x_{i}$ is the input image
        \State Decomposed model forward $y^d_i=F_{d}(x_{i},m_{i})$, where $m_{i}$ is the one-hot task label. Our EKS conv reformulates traditional convolution according to Equ.(\ref{eks})
        \If {Task==segmentation}
        \State Compute loss function according to Equ.(\ref{loss-seg})
        \ElsIf {Task==classification}
        \State Compute loss function according to Equ.(\ref{loss-cls})
        \EndIf
        \State Backward propagation for decomposed model
        \State Return $y^d_i$
        \EndFor
		\State Return $Y^d$
	\end{algorithmic} 
	\caption{\textbf{Low-Rank Knowledge Decomposition~(LoRKD)}: 
    }
	\label{alg:pipeline_usersim}
    \vspace{-0.05cm}
\end{algorithm}

For clarity, we summarize the decomposition procedure of LoRKD in Algorithm~\ref{alg:pipeline_usersim}.
At the beginning of training, we first freeze the low-rank expert modules and train only the backbone of the model.
The introduction of this warmup phase offers two key benefits.
Firstly, the low-rank structure needs to be attached to a well-trained backbone.
Training the backbone first, before integrating the low-rank expert modules, ensures that general and task-specific knowledge are effectively separated.
Secondly, training during the warmup phase provides priors about the difficulty of learning in different regions, providing guidance on how to set the rank of low-rank expert modules in subsequent phases~\S\ref{imbalance-rank}.
After the warmup phase, the low-rank experts are trained together with the shared backbone.

To show the computational merit, we compare our efficient knowledge separation convolution with FLoRA~\cite{wen2023batched},
a recent parameter-efficient fine-tuning method that utilizes multiple low-rank adapters like us. FLoRA allows each example in a minibatch to have its unique low-rank adapters and demonstrates lower computational costs compared to the vanilla manner.
Their comparision \textit{w.r.t.} computational complexities is presented in the following table:
\begin{table}[h]
    \centering
    \vspace{-0.2cm}
    \resizebox{0.48\textwidth}{!}{
    \setlength{\tabcolsep}{1mm}{
    \begin{tabular}{c|c|c}
    \toprule[1.5pt]
        Method & Improved Operation & Computational Cost\\
         \midrule
         FLoRA & $\mathbf{Y}=\mathbf{A}\circ\left((\mathbf{B}\circ\mathbf{X})\mathbf{W_0}\right)$&$c_2(rbld^2)$\\
         \midrule
         EKS conv (ours) & $\mathbf{Y}=\mathbf{X}(\mathbf{W_0}+\sum\nolimits_{i=1}^{T} (\mathbf{\widetilde{BA}}\odot \mathbf{M})_i)$&$Tc_2(rd^2)+c_2(bld^2)$\\
    \bottomrule[1.5pt]
    \end{tabular}}}
    \label{tab:r1q1}
    \vspace{-0.2cm}
\end{table}

\noindent Following the notation in~\cite{wen2023batched}, we omit the cost of element-wise multiplications  (``$\circ$'') and omit the dimensions as $\mathbf{W}\in\mathbb{R}^{d \times k}, \mathbf{A}\in\mathbb{R}^{r \times k}, \mathbf{B}\in\mathbb{R}^{d \times r}$. Here, $c_2$ represents the computational coefficient of matrix multiplication, $b$ is the batch size, $l$ is the sequence length, and $T$ is the number of tasks.  
For EKS conv to be more efficient than FLoRA, the following condition must be satisfied:
\begin{equation}
 \frac{rbld^2}{Td^2r+bd^2l} \ge 1  \Longrightarrow  \frac{Tr}{bl} + 1\le r\nonumber
\end{equation}
This inequality typically holds in most real-world cases, as $bl > Tr$ and $r > 2$ are common training settings. 
The key difference is that while FLoRA reduces costs by replacing expensive batched matmuls (bmm) with element-wise multiplications (``$\circ$'') and broadcasting, our method further reduces computational costs by performing early parameter fusion before the forward pass of DNNs.
In summary, our approach surpasses the efficiency of FLoRA through early parameter fusion. 
Additionally, FLoRA uses broadcasting to improve efficiency, which cannot be well generalized to convolution operations, while LoRKD is not subject to this.

\section{Experiments}  
\label{experiments}

\begin{table}[t]\footnotesize
\centering
\caption{Detailed statistics of the datasets.
}
\label{tab:dataset-all}
\resizebox{0.48\textwidth}{!}{
\setlength{\tabcolsep}{3mm}{
\begin{tabular}{c|c|ccccc}
\toprule[1.5pt]
\multirow{7}{*}{\rotatebox[origin=c]{90}{Segmentation}} & \multirow{2}{*}{Pretraining} & Dataset & Task & Modality & Image & Mask  \\ 
\cmidrule{3-7} 
& & SAT-DS~\cite{zhao2023one} & 8 & 2 & 13303 & 214816 \\
\cmidrule{2-7}
\cmidrule{3-7}
& \multirow{5}{*}{Downstream} & MSD\_Hippocampus~\cite{antonelli2022medical}  &1 & MRI & 260 & 780  \\ 
& & CHAOS\_CT~\cite{kavur2021chaos}  &1 & CT & 20 & 20  \\ 
& & MSD\_Liver~\cite{antonelli2022medical}  &1 & CT & 131 & 262   \\ 
& & COVID19~\cite{ma2021toward} & 1 & CT  & 20 & 80  \\ 
& & MSD\_Spleen~\cite{antonelli2022medical} & 1 & CT  & 41 & 41 \\ 
\midrule[1.5pt]
\multirow{11}{*}{\rotatebox[origin=c]{90}{Classification}} & \multirow{4}{*}{Pretraining} & Dataset & Task & Modality & Label & Image  \\ 
\cmidrule{3-7}
& & Radimagenet~\cite{mei2022radimagenet} & 11 & 3 & 165 & 1354886 \\
& & MedMnist~\cite{yang2023medmnist} & 10 & 8 & 73 & 705689 \\
& & Med-MT & 8 & 5 & 57 & 119655 \\
\cmidrule{2-7} 
\cmidrule{3-7}
& \multirow{7}{*}{Downstream} & COVID~\cite{xingyi2020covid_CT}  &1 & CT  & 2  & 746  \\ 
& & BTC~\cite{saleh2020BTC}  &1 & MRI & 4 & 3538  \\ 
& & AD~\cite{AD}  &1 & MRI & 4 & 3264   \\ 
& & Mura\_shoulder~\cite{rajpurkar2017mura} & 1 & MRI & 2  & 8942   \\ 
& & AUTID~\cite{AUITD}  &1 & Ultrasound  & 3 & 6400 \\ 
& & HAM10000~\cite{tschandl2018ham10000} & 1& Dermatoscope  & 7 &  10015\\ 
& & DET10~\cite{liu2020chestxdet10} &1 &  Xray  & 10 & 3543 \\ 
\bottomrule[1.5pt]
\end{tabular}}}
\vspace{-0.3cm}
\end{table}

In this section, we present the experimental results of knowledge decomposition using LoRKD. 
We evaluate its performance on representative medical foundation models for both segmentation and classification tasks, detailing the experimental setup \S\ref{exp-setup}. 
Extensive experiments on pre-training and downstream datasets validate the generalization and transfer capabilities of the decomposed models \S\ref{decompose}. 
\S\ref{cost} provides a detailed cost analysis to verify the efficiency of knowledge decomposition.
Additionally, we include ablation studies, knowledge disentanglement, and visualizations of the results in \S\ref{exp-further}.

\subsection{Experimental Setup}
\label{exp-setup}

\begin{table*}[t]
\centering
\caption{Region-wise Evaluation. The \textbf{boldface} indicates the best results.
Each column represents the performance of different methods/models for specific tasks.
``Parmas" represents the total number of parameters during training.
}
\label{tab:sat-results}
\vspace{-0.1cm}
\resizebox{0.95\textwidth}{!}{
\setlength{\tabcolsep}{3mm}{
\begin{tabular}{c|c|c|cccccccc|c}
\toprule[1.5pt]
   Metric& Method & Params & Abdomen & Brain & H\&N & LL & Pelvis & Spine & Thorax & UL& Avg \\
\midrule
\multirow{7}{*}{DSC$\uparrow$}
& nnUNet &1545M & \textbf{87.05} &\textbf{81.93} & 72.08& 82.48 &84.68 &\textbf{81.75} & 86.90& 88.54& 83.18 \\
\cmidrule{2-12}
& SAT-Nano &109.19M& 78.18& 74.00&76.74 &76.27 &80.61 &72.44 &80.69 & 84.74 & 77.96 \\
&  LoRKD-Nano &67.01M & 80.06&73.80 &75.15 & 83.69& 89.28& 70.47& 81.86& 82.34& 79.58\\
&  LoRKD*-Nano &67.32M&80.50&73.96&75.65&85.96&88.49&71.38&82.03&82.47&80.05\\
\cmidrule{2-12}
&SAT-Pro &475.56M& 83.16&77.52 &\textbf{79.27}  &81.53 & 88.28& 72.54& 86.50& 86.23& 81.88 \\
&  LoRKD-Pro &129.10M& 80.56&75.79 &78.61 & \textbf{88.56}& 91.75& 73.68& 87.00& 86.69& 82.83\\
&  LoRKD*-Pro &127.86M& 80.81&75.76 &78.74 & 87.93& \textbf{92.07}& 75.37& \textbf{87.72}& \textbf{89.55}& \textbf{83.49}\\
\midrule
\multirow{7}{*}{NSD$\uparrow$}
& nnUNet &1545M & \textbf{79.70} & \textbf{81.96}& 74.18& 80.02 &76.34 &\textbf{77.72}  & 83.55&83.86 &79.67  \\
\cmidrule{2-12}
& SAT-Nano & 109.19M& 67.17& 72.54&82.12 &74.84 &76.05 &69.94 &76.80 & 85.95 & 75.68\\
&  LoRKD-Nano &67.01M& 68.02& 71.50& 79.64& 80.07& 84.72& 67.01& 77.75& 83.34& 76.51\\
&  LoRKD*-Nano &67.32M &68.79&71.73&80.49&84.50&84.06&68.02&78.07&83.35&77.37 \\
\cmidrule{2-12}
& SAT-Pro &475.56M&  73.40& 77.46& \textbf{85.24} &80.83 &85.22 & 70.59& \textbf{83.87}& 88.12&  80.59\\
&  LoRKD-Pro & 129.10M& 70.75& 75.92 & 84.95 & \textbf{88.79}& 88.51& 72.14& 82.69& 87.91& 81.46\\
&  LoRKD*-Pro &127.86M& 71.12& 75.88 & 85.12 & 87.90& \textbf{88.89}& 73.98& 83.44& \textbf{90.76}& \textbf{82.14}\\

\bottomrule[1.5pt] 
\end{tabular}}}
\vspace{-0.3cm}
\end{table*}

\subsubsection{Dataset and Foundation Model}
\label{dataset}

To evaluate the decomposition performance on segmentation tasks, we choose a recent state-of-the-art foundation model, Segment Anything in radiology scans by Text prompts (SAT). 
The SAT models come in two sizes: SAT-Nano and SAT-Pro.
They are trained on the SAT-DS dataset, which is the largest and most comprehensive collection of public 3D medical image segmentation datasets~\cite{zhao2023one}.
Furthermore, to determine the extent to which the decomposed expert models can fully replace foundation models in specific domains, we evaluate the transferability of these expert models on five downstream segmentation datasets.

For the classification task, we choose three medical multi-task datasets of varying scales that are popular for medical image diagnosis pre-training: Radimagenet~\cite{mei2022radimagenet}, MedMnist~\cite{yang2023medmnist}, and Med-MT.
We decompose the foundational models pre-trained on these datasets into 11, 10, and 8 lightweight expert models, respectively. 
In addition, we evaluated the transferability of these expert models on seven downstream datasets. 
Detailed information about these datasets can be found in Table~\ref{tab:dataset-all}.

\subsubsection{Evaluation Metrics}
\label{eva-mertrics}
We use two metrics: Dice Similarity Coefficient (DSC) and Normalized Surface Dice (NSD) to evaluate the performance of segmentation models.
Region-wise results are reported for eight regions of the human body: Brain, Head and Neck, Thorax, Abdomen, Pelvis, Spine, Upper Limb, and Lower Limb.
Specifically, we merge results from all segmentation classes within the same region to indicate the general performance in that region.
The average of all region-wise results represents the overall performance.

For the classification task, we also use the accuracy of each region and the average of all region-wise accuracy to evaluate the classification performance.
The division of regions for each dataset varies according to the data type.

\subsubsection{Baselines}
\label{baseline}
For the segmentation task, we compare our decomposed model with the original foundation model and nnUNet~\cite{isensee2021nnu}, which represent the state-of-the-art universal models and specialist models, respectively.
For nnUNet, we train 49 separate models, each specialized on a different sub-dataset, and report their aggregated results.
This makes nnU-Net a strong baseline, as it is an ensemble of specialist models, each optimized individually on specific sub-datasets.

To ensure a more comprehensive comparison, we implemented various baseline methods on less resource-demanding classification tasks.
The competitive baselines are as follows:
(1) \textbf{Baseline} refers to training from scratch on downstream tasks.
(2) \textbf{Single-Task Learning (STL)} refers to training multiple single-task networks independently, similar to ``nnUNet" in segmentation experiments.
(3) \textbf{Multi-Task Learning (MTL)} refers to training a single model to predict all tasks.
(4) \textbf{STL-KD} and (5) \textbf{MTL-KD} correspond to the KD version of STL and MTL, respectively, which utilize knowledge distillation to transfer knowledge from foundation models.
(6) \textbf{MoCo-MTL}~\cite{fernando2022mitigating} and (7) \textbf{Aligned-MTL}~\cite{senushkin2023independent} are the advanced MTL algorithms.
(8) \textbf{KF}~\cite{yang2022factorizing} represents the advanced knowledge decomposition method, which is the closest to our goal and serves as our primary comparison object in classification.

\subsubsection{Implementation Details}
\label{Implementation}
For both decomposition training and downstream fine-tuning in segmentation experiments, we use AdamW optimizer with a learning rate of 1e-4 and CosineAnnealingLR as the scheduler.
The default values for the hyperparameters are set as follows: $\beta$=0.1, $r$=8.
The vision backbone of all models is based on the 3D U-Net~\cite{ronneberger2015u} structure of varying sizes.
During decomposition, we directly inherit the text encoder from the foundation model and keep it frozen.

For the decomposition training in the classification task, we use the SGD optimizer with a learning rate of 0.05 and CosineAnnealingLR as the scheduler for training 100 epochs. For the downstream fine-tuning, we use AdamW optimizer with a learning rate of 5e-5 and train the model for 240 epochs. 
The default values for the hyperparameters are set as follows: $\beta$=1, $r$=8. The pre-trained model structure is ResNet50~\cite{he2016deep}, and the structure of the lightweight decomposition model is ShuffleNetV2~\cite{ma2018shufflenet}. 

\subsection{Main Results}
\label{decompose}

\subsubsection{Performance in Segmentation}
\paragraph{Decomposition Performance in Segmentation}

The region-wise evaluation results are shown in Table~\ref{tab:sat-results}.
Each column corresponds to a specific region.  
``Parmas" represents the total number of parameters during training.
We use ``-nano" and ``-pro" to distinguish between two sizes of models and use ``*" to distinguish between LoRKD-balance and LoRKD-imbalance.

\begin{table*}[th]\footnotesize
\centering
\caption{The transfer performance of the decomposed expert models on six downstream segmentation datasets. 
}
\label{tab:sat-downstream}
\vspace{-0.2cm}
\resizebox{0.9\textwidth}{!}{
\setlength{\tabcolsep}{3mm}{
\begin{tabular}{c|c|ccccc|c}
\toprule[1.5pt]
   Metric& Method & Hippocampus & Liver & COVID19 & Spleen & CHAOS\_CT & Avg \\
\midrule
\multirow{7}{*}{NSD$\uparrow$}
& nnUNet &\textbf{97.92}  &63.78 &77.02 &\textbf{88.01} &81.04  &81.55 \\
\cmidrule{2-8}
& SAT-Nano & 95.60 & 52.89 & 71.18 &80.43 & 81.16 & 76.25 \\
&  LoRKD-Nano & 96.44& 62.96&  76.34& 84.59& 85.75& 81.22\\
&  LoRKD*-Nano & 96.59& 63.40 &  77.74 & 86.33 & \textbf{86.13} & 82.04\\
\cmidrule{2-8}
& SAT-Pro & 96.45 &62.89 & 72.82 &84.86 & 84.63 &80.33 \\
&  LoRKD-Pro & 96.75 & \textbf{65.73} &  79.27 & 86.49 & 85.65 & 82.78 \\
&  LoRKD*-Pro & 96.62 & 65.01 & \textbf{79.43} & 87.47 & 85.91 & \textbf{82.89}  \\
\midrule
\multirow{7}{*}{DSC$\uparrow$}
& nnUNet & \textbf{89.18} &77.92 &\textbf{91.53}  & 92.95& 97.08 &\textbf{89.73} \\
\cmidrule{2-8}
& SAT-Nano & 86.20 &68.46 & 82.57 &93.49 &96.55  &85.46 \\
&  LoRKD-Nano & 87.51 & 75.71 & 86.10 & 93.96 & 97.17 & 88.09  \\
&  LoRKD*-Nano & 87.56 & 76.03 &  87.47 & 94.41 & \textbf{97.26} & 88.55  \\
\cmidrule{2-8}
& SAT-Pro & 87.62 &76.63 &83.18  &94.12 &97.02  &87.72 \\
&  LoRKD-Pro & 87.65 & \textbf{78.10} &  88.74 & 94.51 & 97.18 & 89.23  \\
&  LoRKD*-Pro & 87.90 & 77.66 &  89.01 & \textbf{94.65} & 97.27 & 89.30  \\

\bottomrule[1.5pt] 
\end{tabular}}}
\vspace{-0.3cm}
\end{table*}

\par
\begin{figure*}[t]
    \centering
    \subfigure[Cost comparison on segmentation]{
        \centering
        \includegraphics[width=0.48\textwidth]{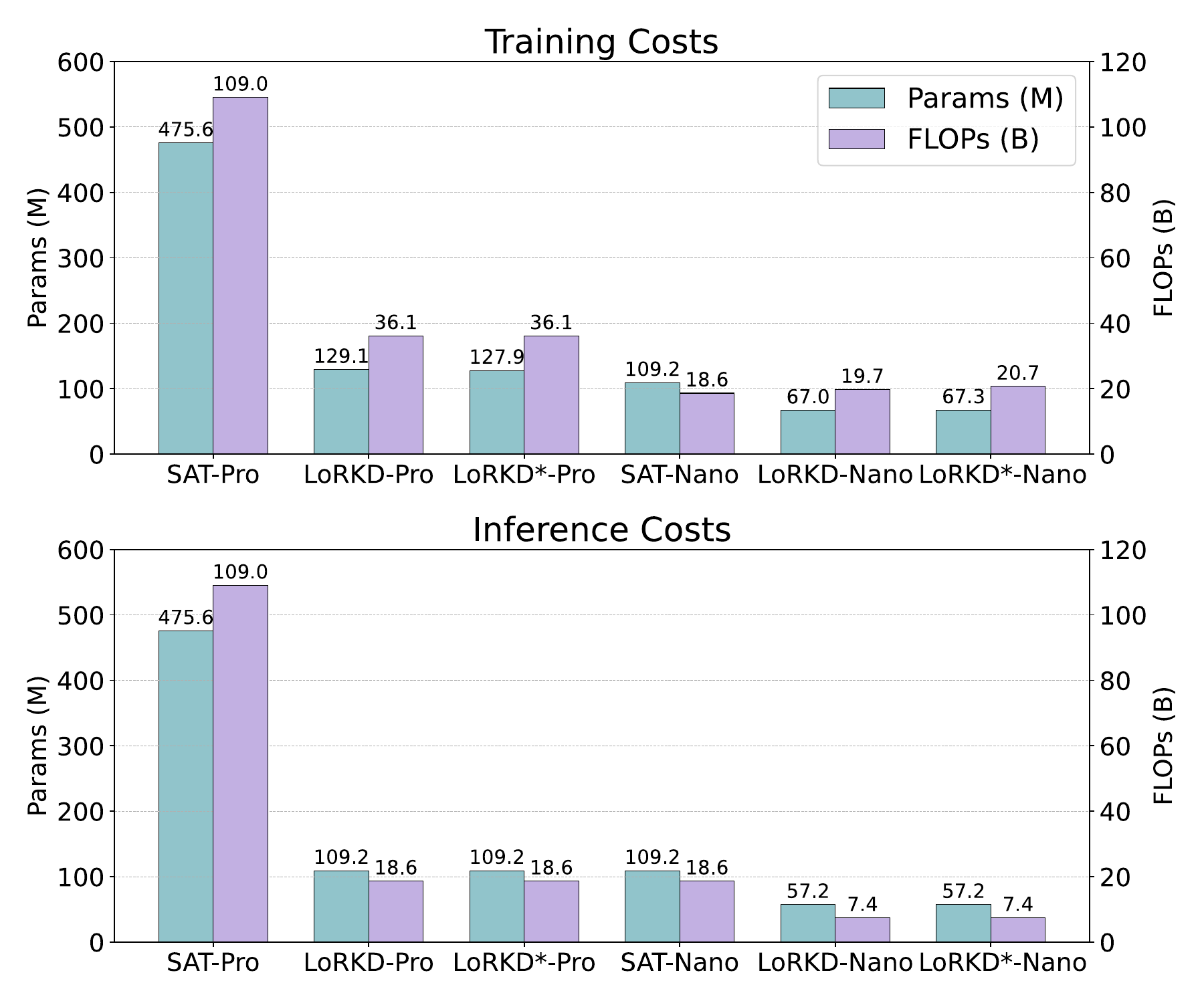}
        
        \label{fig:cost-seg}
    }
    \subfigure[Cost comparison on classification]{
        \centering
        \includegraphics[width=0.48\textwidth]{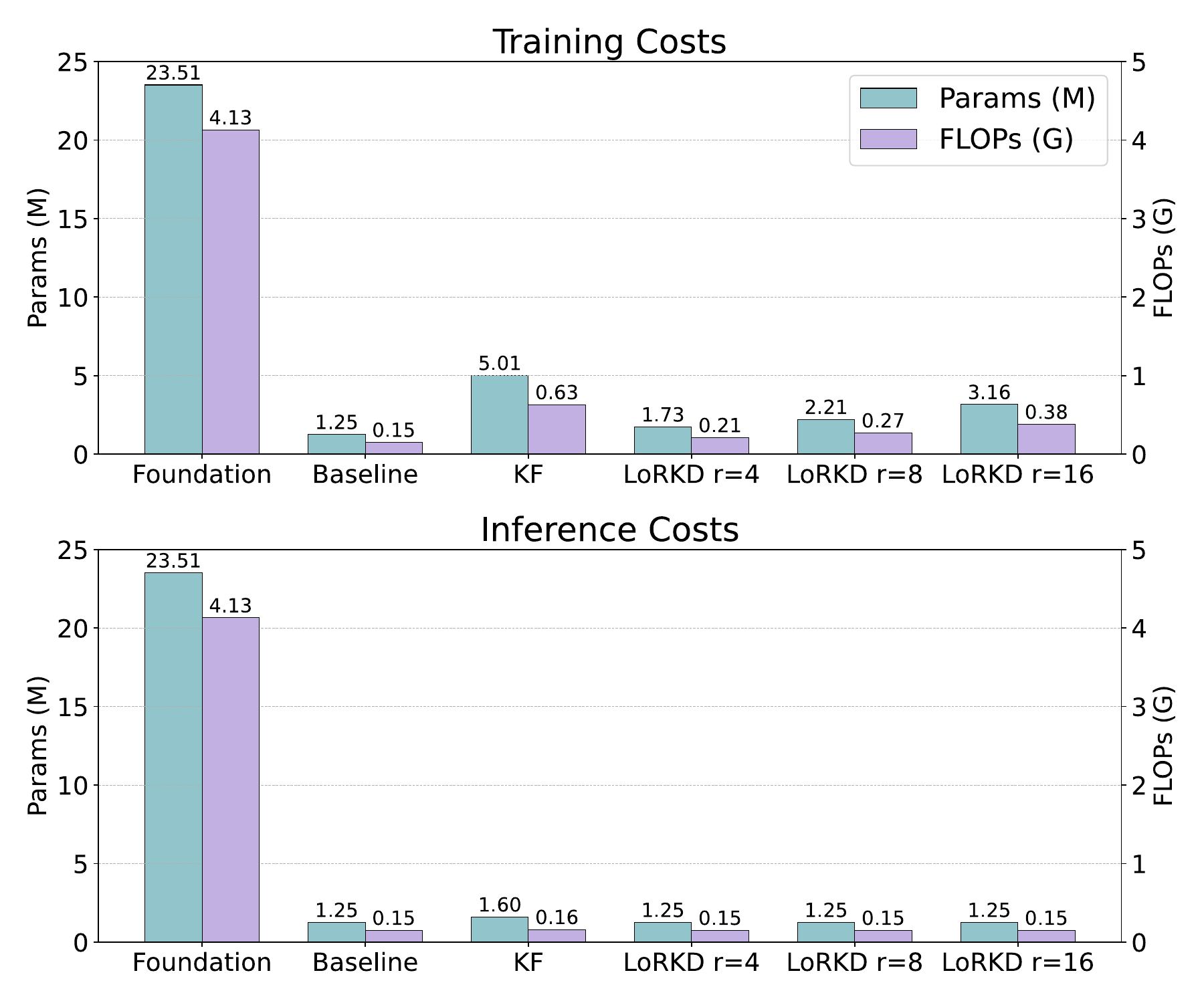}
        \label{fig:cost-cls}
    }
    \caption{Cost comparison between different models.
    We calculate the resource consumption in both training and deployment scenarios.
    Fig~\ref{fig:cost-seg} shows the results of the segmentation task, while Fig~\ref{fig:cost-cls} displays the classification results.
    }
    \vspace{-0.3cm}
    \label{fig:cost}
\end{figure*}

{\noindent \bf Decomposed model vs. Foundation model.}
In general, the decomposed model can achieve stronger specialization with lower costs
(The cost comparison can be found in Figure~\ref{fig:cost}).
For SAT-Pro, our decomposed model has only 23\% of its parameters and 17\% of its computational overhead, yet it surpasses the foundation model with approximately a 2\% performance improvement.
Similarly, for SAT-Nano, our decomposed model has 52\% of its parameters and 40\% of its computational overhead, and it also outperforms SAT-Nano on both metrics.
Notably, in four regions, \shortname provides considerable performance gains, up to 8\%.
In the other regions, \shortname can also maintain performance comparable to the foundation model with fewer parameters.
This demonstrates that our method not only achieves lossless decomposition but also surpasses the original model by alleviating the conflict between heterogeneous tasks.

\par{\noindent \bf LoRKD* vs. LoRKD.}
As shown in Table~\ref{tab:sat-results}, LoRKD* consistently outperforms LoRKD in most cases, demonstrating the effectiveness of our imbalanced rank design.
This indicates that the loss reduction in the warmup phase can accurately reflect the learning difficulty of each region, thereby guiding the reasonable allocation of parameters for each region.
In detail, LoRKD*-Pro exhibits higher DSC scores than LoRKD-Pro in 7 out of the 8 regions—except the Lower Limb.
Similarly, LoRKD*-Nano outperforms LoRKD-Nano in most regions, except for the Pelvis.
The regions where LoRKD*-Nano and LoRKD*-Pro perform worse differ because their backbone models have different sizes, leading to varying learning capabilities in each region.
Consequently, the automatically computed rank values of each low-rank expert module differ between LoRKD*-Nano and LoRKD*-Pro.

\par{\noindent \bf Decomposed model vs. Ensemble of SOTA Specialist model.}
It is worth noting that our LoRKD*-Pro can surpass nnUNet in overall performance (``Avg"), filling the performance gap between universal models and specialist models.
This is particularly challenging since nnUNet represents an ensemble of 49 state-of-the-art models trained independently on each sub-dataset.
Specifically, in the five regions of Head \& Neck, Upper Limb, Lower Limb, Pelvis, and Thorax, LoRKD*-Pro consistently outperforms nnUNet.
This indicates that the tasks in these regions benefit from universal training, and all tasks within these regions can be addressed by a single expert model.
However, in the three regions of the Abdomen, Brain, and Spine, LoRKD*-Pro remains inferior to the nnUNet ensemble.
This suggests that these regions are suitable for fine-grained specialist models, as universal models still struggle to adequately solve the tasks in these regions.

\paragraph{Transfer Performance in Segmentation}

For the decomposed lightweight expert model to fully replace the foundation model in a specific domain, it is essential that the expert models not only perform well on the same distribution of data (pre-training dataset) but also demonstrate their generalization ability on downstream tasks with similar distributions.
Hence, we evaluate the performance of the decomposed model and baselines on several representative downstream datasets.


Table~\ref{tab:sat-downstream} presents the performance comparison between the decomposed expert models and the baselines on five downstream segmentation datasets.
For the specialist nnUNet, we directly train five models on each downstream dataset.
As for the decomposed model, we fine-tune the expert model corresponding to the downstream dataset, such as using the brain expert for the MSD\_Hippocampus dataset.
As for the foundation model, we fine-tune the pre-trained model on the downstream dataset.

\par{\noindent \bf Foundation model vs. Specialist model.}
It can be observed that the overall performance of nnUNet on downstream datasets exceeds that of the foundation model.
nnUNet demonstrates superior average performance, surpassing SAT in 4 out of 5 datasets, with the only exception being the CHAOS\_CT dataset.
This indicates that despite the universal pre-training knowledge of foundation models, their ability to transfer to downstream data is insufficient to replace specialist models.
Downstream datasets typically have only a small amount of data, which makes them difficult to support the fine-tuning of the foundation model with numerous parameters, according to the Scaling Law~\cite{kaplan2020scaling}.

\begin{table*}[t]\footnotesize
\caption{The decomposition performance on classification pre-training datasets.  
It is worth noting that except for KF and ours, the concept of knowledge decomposition does not exist in other methods. 
The presence of homonymous experts implies different modalities. 
For more details, please refer to the supplementary materials.}
\vspace{-0.25cm}
\centering
\resizebox{\textwidth}{!}{
\setlength{\tabcolsep}{2mm}{
\begin{tabular}{c|c|ccccccccccc|c}
\multicolumn{14}{c}{\textbf{Radimagenet (1.35 million images, 11 tasks)}} \\
\toprule[1.5pt]
Method  &{Params(M)} & Lung & Abdomen & Thyroid & Abdomen & Knee & Shoulder & Spine & Ankle & Abdomen  & Brain & Hip  & Avg \\
\midrule
{Foundation} & 23.51  & 36.22 & 46.52 & 74.05 & 48.42 & 40.09 & 31.32 & 17.79  & 12.95 & 64.17 & 77.30 & 32.33   & 43.74\\
\midrule
{STL} & 13.79 & 76.42 & 33.94 & 91.55 & \textbf{69.17} & \textbf{49.32} & 41.80 & 20.62 & \textbf{20.31} & 65.99 & 83.88 & 51.05  & 54.91\\
{MTL} & 1.25 & 77.16 & 37.45 & 91.73 & 68.43 & 46.47 & 42.72 & 20.85 & 18.17 & 71.13 & \textbf{84.67} & 55.16 & 55.81 \\
{STL-KD} & 13.79 & 78.00 & 31.74 & 91.34 & 69.10 & 46.57 & 43.09 & 19.77 & 19.43 & 69.85 & 83.83 & 52.19   & 54.99\\
{MTL-KD} & 1.25 & \textbf{78.92} & 33.89 & 91.97 & 68.54 & 48.51 & 43.34 & 21.03 & 18.48 & 69.58 & 84.18 & 54.90  & 55.75\\
MoCo-MTL & 1.25 & 76.28 & \textbf{45.56} & 86.26 & 67.00 & 45.58 & \textbf{43.97} & 18.74 & 17.41 & \textbf{74.88} & 84.33 & 52.71  & 55.70\\
Aligned-MTL & 1.25 & 77.74 & 36.38 & 91.76 & 68.51 & 48.41 & 43.28 & 21.26 & 18.37 & 68.57 & 84.54 & 54.86  & 55.79\\
\midrule
{KF} & 5.01 & 64.57 & 20.38 & \textbf{95.82} & 68.05 & 45.56 & 39.03 & \textbf{24.18} & 16.69 &56.65 & 78.46 & 51.74  & 51.01\\
{LoRKD} & 2.21 & 78.72 & 36.95 & 91.87& 68.77 & 48.80 & 43.26 & 21.41 & 19.26  & 69.24 & 84.60 & \textbf{55.93}  & \textbf{56.26}  \\
\bottomrule
\end{tabular}}}
\label{upstream}
\vspace{-0.15cm}
\end{table*}
\begin{table*}[ht!]\footnotesize
\vspace{-0.2cm}
\centering
\resizebox{\textwidth}{!}{
\setlength{\tabcolsep}{3mm}{
\begin{tabular}{c|c|cccccccccc|c}
\multicolumn{13}{c}{\textbf{MedMnist (705,689 images, 10 tasks)}} \\
\toprule[1.5pt]
Method &{Params(M)}  & Colon & Retinal & OrganC & Cell & Breast & Tissue & Skin & OrganA & OrganS & Chest   & Avg \\
\midrule
Foundation & 23.51 & 87.41 & 77.40 & 23.51 & 50.37  & 84.62 & 40.55 & 12.92 & 18.64 & 18.90 & 86.22   & 50.05\\
\midrule
STL& 12.54 & 84.53 & 78.40 & 89.65 & \textbf{96.81} & 85.26 & \textbf{68.89} & 73.97 & 92.90 & 77.43 & 85.42 &    83.33\\
MTL& 1.25 & 80.99 & 77.10 & 89.90 & 95.67 & 83.33 & 65.42 & 74.21 & 91.33 & 76.34 & 86.89 &     82.12\\
STL-KD& 12.54 & 84.33 & 77.10 & 90.45 & 96.52 & 83.33 & 68.25 & \textbf{74.81} & \textbf{93.53} & \textbf{77.52} & 82.53 &    82.84\\
MTL-KD& 1.25 & 82.83 & 75.20 & 90.02 & 95.94 & 83.26 & 64.56 & 74.31 & 92.13 & 76.02 & 86.39 &     82.06\\
MoCo-MTL& 1.25 & 76.10 & 69.80 & 80.00 & 86.55 & 76.92 & 63.89 & 69.18 & 83.82 & 67.81 & 83.87 &    75.79\\
Aligned-MTL& 1.25 & 79.78 & 73.10 & 89.70 & 95.44 & \textbf{88.46} & 64.00 & 74.36 & 90.81 & 75.06 & 86.22 &   81.69 \\
\midrule
{KF}& 4.67  & 37.83 & 48.20 & 72.40 & 44.93 & 80.13 & 54.17 & 38.01 & 71.75 & 59.19 & 72.12   & 57.87\\
{\textbf{LoRKD}}& 2.12  & \textbf{83.90} & \textbf{78.60} & \textbf{90.57} & 96.26 & 87.18 & 67.01 & 73.97  & 92.83 & 77.27 & \textbf{87.39}   & \textbf{83.50} \\
\bottomrule
\end{tabular}}}
\vspace{-0.15cm}
\end{table*}
\begin{table*}[ht!]\footnotesize
\vspace{-0.2cm}
\centering
\resizebox{\textwidth}{!}{
\setlength{\tabcolsep}{4.2mm}{
\begin{tabular}{c|c|cccccccc|c}
\multicolumn{11}{c}{\textbf{Med-MT (119,655 images, 8 tasks)}} \\
\toprule[1.5pt]
Method &{Params(M)}  & Retinal & Skin & Breast & GI tract & Lung & Shoulder & Lung & Bone    & Avg\\
\midrule
{Foundation}& 23.51 & 81.83 & 87.01 & 81.82 & 91.25 & 66.37 & 92.31 & 65.00 & 59.46    &78.13\\
\midrule
{STL}& 10.03  & 75.27 & 77.92 & 76.59 & 85.62 & 69.91 & 75.00 & 64.85 & 51.15 &   72.04\\
{MTL}& 1.25 & 78.14 & 78.57 & 77.85 & 87.94 & 69.91 & 79.81 & 64.37 & 49.41     & 73.25\\
{STL-KD}& 10.03  & 71.45 & 67.53 & 77.18 & 86.06 & 60.18 & 78.85 & 64.67 & 51.23    & 69.64 \\
{MTL-KD}& 1.25  & 79.23 & 77.27 & 77.89 & 88.06 & 76.11 & 77.88 & 64.84 & 49.17    & 73.80 \\
MoCo-MTL& 1.25  & 58.74 & 55.84 & 51.74 & 48.31 & 67.26 & 67.31 & 46.76 & 20.11     & 52.01\\
Aligned-MTL& 1.25  & 61.07 & 56.49 & 51.50 & 52.63 & 69.03 & 67.31 & 46.77 & 19.17     & 53.00\\
\midrule
{KF} & 3.99 & 65.30 & 74.67 & 52.19 & 61.12 & \textbf{77.88} & 79.81 & 60.21 & 33.50     & 63.09\\
{\textbf{LoRKD}} & 1.95 & \textbf{79.37} & \textbf{85.06} & \textbf{79.04} & \textbf{88.63} & 72.57 & \textbf{83.65}& \textbf{65.07}  & \textbf{52.42}     & \textbf{75.73}\\
\bottomrule[1.5pt]
\end{tabular}}}
\vspace{-0.4cm}
\end{table*}

\par{\noindent \bf Decomposed model vs. Baselines.}
Generally, our decomposed models yield favorable results and significantly surpass the original foundation models.
Compared to SAT-Nano, LoRKD*-Nano demonstrates a 5.8\% performance improvement on NSD and a 3.1\% improvement on DSC.
For SAT-Pro, LoRKD*-Pro achieves a 2.6\% performance improvement on NSD and a 1.6\% improvement on DSC.
Notably, the performance of LoRKD*-Nano and LoRKD-Nano is comparable to or better than the larger model SAT-Pro, indicating that compact expert models are more suitable for downstream datasets than the foundation model.
Compared to nnUNet, LoRKD*-Pro and LoRKD-Pro achieve comparable performance, outperforming nnUNet in three out of five datasets.
We also observe that LoRKD-Pro consistently outperforms LoRKD-Nano, and LoRKD* is slightly better than LoRKD.
This demonstrates that the performance on the pre-training dataset is positively correlated with the transferability to downstream tasks.
Models that perform better on the pre-training tasks tend to exhibit superior transferability to downstream tasks.

\subsubsection{Performance in Classification}

\paragraph{Decomposition Performance in Classification}

The performance comparison of different methods on three pre-training classification datasets is presented in Table~\ref{upstream}.
Each column corresponds to a specific task. Only KF and our method focus on the knowledge decomposition of pre-trained models.
Considering the generalization requirements of foundation models, it is typical for these models to employ a unified classification head during training rather than configuring a specific classification head for each task~\cite{mei2022radimagenet}.
This practice accounts for the relatively poor performance of the foundation model depicted in Table~\ref{upstream}.

\par{\noindent \bf The foundation model vs. STL.} 
The performance of the foundation model surpasses that of STL on the Med-MT dataset but is significantly inferior to STL on both Radimagenet and MedMnist, especially MedMnist. This observation suggests that as the scale and diversity of the pre-training dataset increase, the specialization of the pre-trained model gradually diminishes due to conflicts between different domain knowledge. In contrast, training models independently for each task (STL) can prevent interference between different tasks, resulting in superior performance on Radimagenet and MedMnist compared to foundation models. However, STL cannot learn common knowledge across tasks, often necessitating more data to ensure generalization. Additionally, training 
$T$ individual models is not only time-consuming but also leads to a linear increase in the number of parameters.

\begin{table*}[!th]\footnotesize
\caption{The transfer performance of the decomposed expert models on seven downstream classification datasets. ``Comp. Ratio" denotes the compression ratio, defined as the ratio of the deployed model parameters to the parameters of the foundation model. ``-" indicates the absence of data corresponding to the downstream tasks in the pre-training dataset.}
\vspace{-0.1cm}
\centering
\resizebox{0.98\textwidth}{!}{
\begin{tabular}{c|c|cc|c|c|c|c|c|c|c|c}
\toprule[1.5pt]
{Pre-train} & {Model}& {Params} & {Comp. Ratio} & {COVID} & {BTC} & {AD} & {Mura\_s} & {AUITD} & {HAM10000} & {DET10}& {Avg}  \\
\midrule
\multirow{11}{*}{\rotatebox[origin=c]{90}{Radimagenet}} &Foundation  & 23.51M & / & 78.33 & 80.20 & 74.35 & 71.05 & 96.66 & 75.08 & 86.69 & 80.34 \\
\cmidrule(r){2-12} 
&Baseline  & 1.25M & 5.32\% & 82.76 & 75.38 & 76.08& 76.73 & 97.77 & 74.42  & 87.54 & 81.52 \\
&STL & 1.25M & 5.32\% & 82.76 & 78.93 & 76.70 & 77.26 & 97.77 & - & 87.52 & - \\
&MTL & 1.25M & 5.32\% & 83.25 & 79.95 & 74.67 & 76.91 & 97.77 & 75.83 & 86.82 & 82.17 \\
&STL-KD & 1.25M & 5.32\% & 82.27 & 80.46 & 76.31 & 76.73 & 96.66 & - & 87.25 & - \\
&MTL-KD & 1.25M & 5.32\% & 81.77 & 78.93 & 73.89 & 76.55 & 96.66 & 74.37 & 87.17 & 81.33 \\
&MoCo-MTL & 1.25M & 5.32\% & 78.82 & 78.68 & 69.27 & 75.49& 91.64 & 71.77  & 86.43 & 78.87 \\
&Aligned-MTL & 1.25M & 5.32\% & 82.27 & 78.43 & 70.29 & 76.91 & 88.58 & 73.07 & 86.91 & 79.49 \\
\cmidrule(r){2-12} 
&KF & 1.60M & 6.81\% & 80.79 & 79.70 & 71.23 & 74.96 & 96.66 & $74.12^{\dagger}$ & 87.17 & 80.66 \\
&\textbf{LoRKD} & 1.25M & 5.32\% & \textbf{86.21} & \textbf{81.47} & \textbf{79.12} & \textbf{79.57} & \textbf{98.33} & $\textbf{76.03}^{\dagger}$ & \textbf{88.50} & \textbf{84.18} \\
\midrule
\midrule
\multirow{11}{*}{\rotatebox[origin=c]{90}{MedMnist}}  &Foundation  & 23.51M & / & 80.30 & 77.41 & 72.09 & 76.38 & 88.86 & 72.12 & 86.80 & 79.14 \\
\cmidrule(r){2-12} 
&Baseline  & 1.25M & 5.32\% & 82.76 & 75.38 & 76.08& 76.73 & 97.77 & 74.42  & 87.54 & 81.52 \\
&STL & 1.25M & 5.32\% & 83.25 & - & - & - & 97.77 & 71.82 & 87.56 & - \\
&MTL & 1.25M & 5.32\% & 81.28 & 78.68 & 77.17 & 76.19 & 97.77 & \textbf{74.82} & 87.36 & 81.90 \\
&STL-KD & 1.25M & 5.32\% & 79.80 & - & - & - & 97.49 & 73.87 & 86.93 & - \\
&MTL-KD & 1.25M & 5.32\% & 80.79 & 78.62 & 76.62& 75.84 & 98.05 & 73.87 & 87.23 & 81.57  \\
&MoCo-MTL & 1.25M & 5.32\% & 78.82 & 77.16 & 77.80 & 74.95 & 97.77 & 72.77 & 86.82 & 80.87 \\
&Aligned-MTL & 1.25M & 5.32\% & 82.27 & 77.42 & 77.72 & 76.90 & 96.37 & 73.87 & 86.97 & 81.65 \\
\cmidrule(r){2-12} 
&KF & 1.60M & 6.81\% & 80.79 & $77.15^{\dagger}$ & $72.71^{\dagger}$ & $74.77^{\dagger}$ & 96.10 & 72.97 & 87.41 & 80.27 \\
&\textbf{LoRKD} & 1.25M & 5.32\% & \textbf{84.24} & $\textbf{79.70}^{\dagger}$ & $\textbf{79.05}^{\dagger}$ & $\textbf{77.80}^{\dagger}$ & \textbf{98.33} & \textbf{74.82} & \textbf{87.60} & \textbf{83.08} \\
\midrule
\midrule
\multirow{11}{*}{\rotatebox[origin=c]{90}{Med-MT}}  &Foundation  & 23.51M & / & 82.76 & 78.17 & 69.19 & 71.76 & 89.69 & 75.53  & 86.69 & 79.11 \\
\cmidrule(r){2-12} 
&Baseline  & 1.25M & 5.32\% & 82.76 & 75.38 & 76.08& 76.73 & 97.77 & 74.42  & 87.54 & 81.52 \\
&STL & 1.25M & 5.32\% & - & - & - & - & - & 73.77 & - & - \\
&MTL & 1.25M & 5.32\% & 82.76 & 76.65 & \textbf{77.48} &77.09 & 97.49 & 74.92 & 87.15 & 81.93 \\
&STL-KD & 1.25M & 5.32\% & - & - & - & - & - & 74.42 & - & - \\
&MTL-KD & 1.25M & 5.32\% & 82.76 & 75.89 & 74.43 & 76.91 & 97.77 & 74.32 & 87.34 & 81.34 \\
&MoCo-MTL & 1.25M & 5.32\% & 80.79 & 76.40 & \textbf{77.48} & 76.91 & 97.49 & 72.62 & 86.91 & 81.23 \\
&Aligned-MTL & 1.25M & 5.32\% & 79.80 & 75.63 & 76.62 & 76.73 & 97.77 & 73.72 & 87.19 & 81.06 \\
\cmidrule(r){2-12} 
&KF & 1.60M & 6.81\% & $80.79^{\dagger}$ & $74.87^{\dagger}$ & $75.76^{\dagger}$ & $76.73^{\dagger}$ & $98.05^{\dagger}$ & 73.92 & $87.39^{\dagger}$ & 81.07 \\
&\textbf{LoRKD} & 1.25M & 5.32\% & $\textbf{83.25}^{\dagger}$ & $\textbf{77.66}^{\dagger}$ & $76.94^{\dagger}$ & $\textbf{78.33}^{\dagger}$ & $\textbf{98.33}^{\dagger}$ & \textbf{75.18} & $\textbf{87.84}^{\dagger}$ & \textbf{82.50} \\
\bottomrule[1.5pt]
\end{tabular}}
\label{downstream}
\vspace{-0.1cm}
\end{table*}

\par{\noindent \bf MTL-based methods vs. STL-based methods.} 
It can also be observed that MTL outperforms STL on Radimagenet and Med-MT, while underperforming STL on MedMnist. This discrepancy may be attributed to the degree of correlation between tasks within the pre-training dataset, with MedMnist having the most diverse modalities (refer to supplementary materials). Unlike standard MTL, advanced MTL methods such as MoCo-MTL and Aligned-MTL do not yield improvements and may even exhibit worse performance. This suggests that balancing multiple optimization objectives to obtain a better shared encoder is not an effective solution when there are significant differences among tasks.
The knowledge distillation variants of STL and MTL (STL-KD and MTL-KD) do not show significant performance improvement, which suggests that the general features extracted by foundation models offer limited benefits for specific tasks and indirectly reflect the importance of specialized features. It aligns with the design philosophy of our LoRKD.

\par{\noindent \bf LoRKD vs. KF and other methods.} 
Compared to the knowledge decomposition method KF, our approach demonstrates significant performance improvements while introducing fewer parameters. 
Specifically, even with 11, 10, or 8 experts, our method employs less than half the number of parameters used by KF. 
This outcome validates the effectiveness of our low-rank expert modules and the efficient knowledge separation convolution. 
Furthermore, our method achieves the best average performance compared to other non-knowledge decomposition baselines, underscoring the potential of knowledge decomposition in extracting task-specific knowledge.

\paragraph{Transfer Performance in Classification}

The performance comparison of the expert models decomposed from three pre-training datasets on seven downstream classification datasets is shown in Table~\ref{downstream}. 
For KF and our method, we fine-tune the corresponding expert models on downstream datasets, such as using the lung expert model for the COVID dataset. 
In the absence of a corresponding expert model, we fine-tune on the shared backbone, similar to~\cite{yang2022factorizing} (denoted with $^{\dagger}$).
As for other non-knowledge decomposition methods, we use the models trained on the pre-training dataset for fine-tuning to demonstrate the advantages of knowledge decomposition in terms of transferability. 
Please refer to the supplementary materials for further details.

The performance of fine-tuning foundation models is observed to be inferior to the Baseline, reinforcing that foundation models cannot replace task-specific models due to their lack of specialization.
Compared to the Baseline, both STL-based and MTL-based methods show minimal improvement, indicating that focusing solely on task-specific or common knowledge does not enhance transferability.
Conversely, our expert models incorporate both common knowledge and task-specific knowledge, which exhibit strong transferability and even significantly outperform KF.
Another advantage over KF is that our method supports parameter fusion and does not require the simultaneous deployment of two networks (CKN and the corresponding TSN need to be deployed simultaneously in KF).

Furthermore, an interesting phenomenon was observed. In comparison to MTL-KD, our method exhibits significantly better performance on downstream datasets. 
This demonstrates the advantage of knowledge decomposition in transferability, which can not be directly reflected through the decomposition performance.
As the scale of the pre-training dataset increases, the transferability of our decomposed expert models also improves, indicating that increasing the scale of pre-training datasets benefits the transferability of the decomposed model.



\subsection{Efficiency}
\label{cost}
The goal of knowledge decomposition is to break down the foundation model into lightweight expert models. 
These expert models need to be compact enough to ensure higher practicality and deployability. 
Therefore, we have conducted a comprehensive analysis of the model's resource requirements.
We measured the model parameters and computational overhead (FLOPs) during both training and inference stages.
\par{\noindent \bf Lower Costs on Segmentation.}
As shown in Figure~\ref{fig:cost-seg}, our method significantly reduces the resource consumption of the foundation model, indicating that \shortname can effectively lower deployment costs while maintaining high computational efficiency.
For SAT-Pro, the decomposed models can achieve compression ratios of 22.96\% in parameters and 17.09\% in computation.
While for SAT-Nano, the decomposed model achieved compression ratios of 52.42\% in parameters and 39.83\% in computation.
This is highly valuable for deploying the models in real-world scenarios that are resource-constrained in under-developed area.



\begin{figure}[t]
    \centering
    \includegraphics[width=0.98\linewidth]{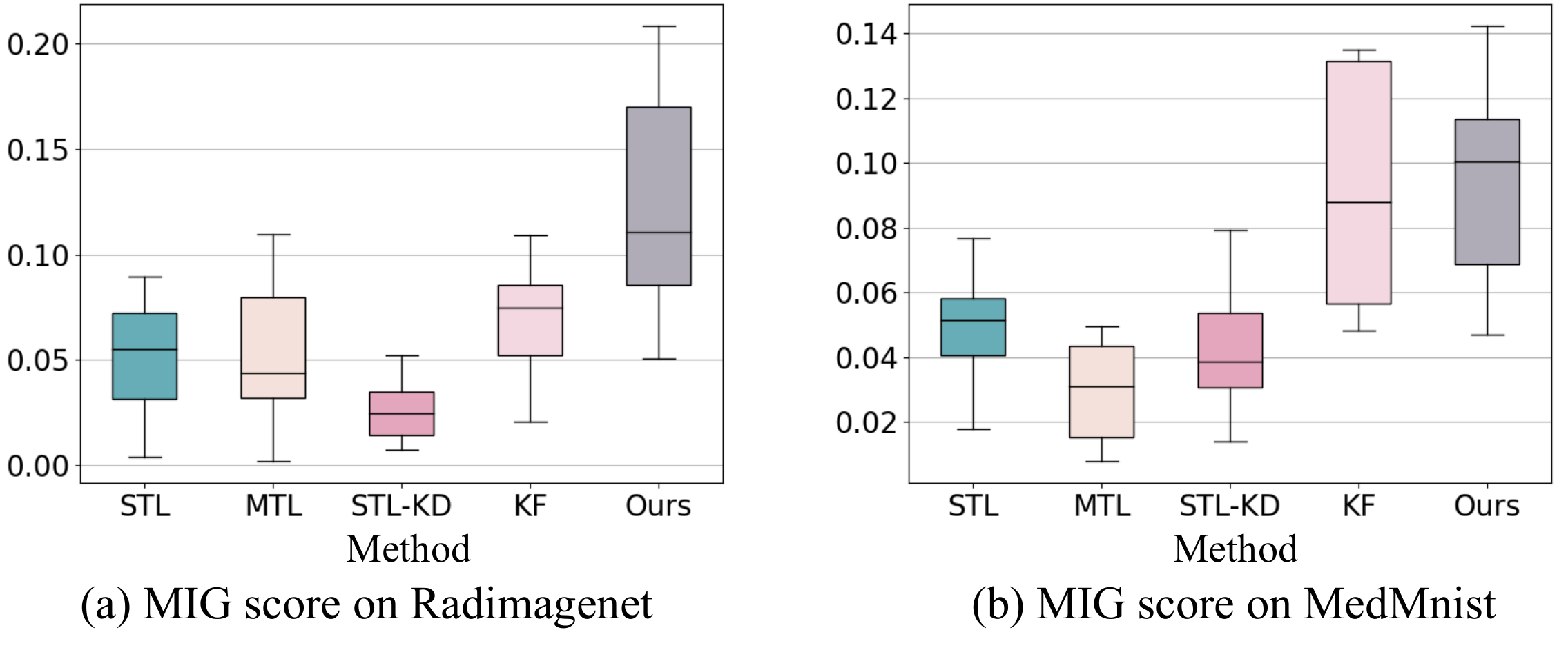}
    \vspace{-0.32cm}
    \caption{The comparison of MIG scores on different methods.}
    \label{fig:mig}
    \vspace{-0.3cm}
\end{figure}
\begin{figure}[t]
    \centering
    \includegraphics[width=\linewidth]{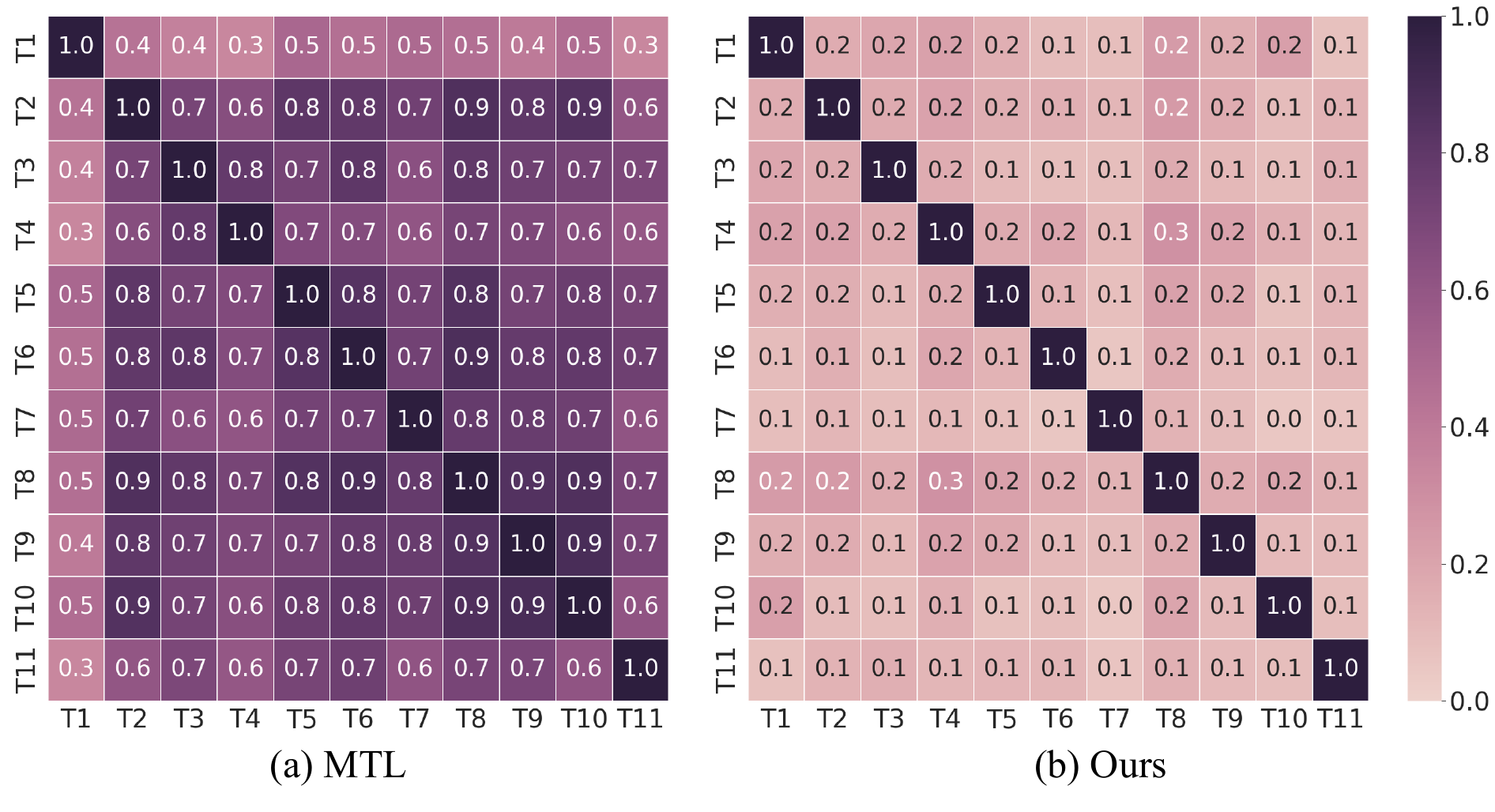}
    \vspace{-0.7cm}
    \caption{The CKA similarity matrices of MTL and \shortname.}
    \label{fig:cka}
    \vspace{-0.5cm}
\end{figure}

\par{\noindent \bf Lower Costs on Classification.}
For the classification task, we compare the costs among different methods on Radimagenet, as shown in Figure~\ref{fig:cost-cls}.
Similar to the segmentation task results, our decomposed models significantly reduce the number of parameters and FLOPs compared to the foundation model.
It is worth mentioning that if parameter fusion is used, our costs will be the same as baseline, achieving a compression ratio of 5.3\% in parameters and 3.6\% in computation.
As $r$ increases, our costs remain minimal and do not increase significantly.
In comparison to KF, even at $r$=16, our method still incurs significantly lower costs.

\subsection{Further Analysis}
\label{exp-further}

\begin{table*}[t]\footnotesize
\centering
\caption{Ablation on LoRA rank on the pre-training segmentation dataset.
}
\label{tab:ablation-r}
\vspace{-0.2cm}
\resizebox{0.95\textwidth}{!}{
\setlength{\tabcolsep}{3mm}{
\begin{tabular}{c|c|c|c|cccccccc|c}
\toprule[1.5pt]
   Metric& Model & Rank & Params & Abdomen & Brain & H\&N & LL & Pelvis & Spine & Thorax & UL& Avg \\
\midrule
\multirow{6}{*}{NSD$\uparrow$}& \multirow{3}{*}{ LoRKD-Nano}& 4 &62.12M &67.85 &71.37 &79.59 & 80.46& 84.66&66.71 & 76.93&78.64 &75.78 \\
& & 8 &67.01M& 68.02& 71.50& 79.64& 80.07& 84.72& 67.01& 77.75& 83.34& 76.51 \\
& & 16 &76.78M &68.03 &71.48 &79.65 &79.58 &84.52 &67.04 &77.28 &81.25 &76.10 \\
\cmidrule{2-13}
& \multirow{3}{*}{ LoRKD-Pro}& 4 &119.15M& 71.11&75.95 &85.09 &86.82 &88.38 &73.06 & 83.55&82.93 &80.86 \\
& & 8 & 127.86M& 70.75& 75.92 & 84.95 & 88.79& 88.51& 72.14& 82.69& 87.91& 81.46\\
& & 16 &149.02M&71.08 &75.86 &85.13 &85.53 &89.06 &72.12 &83.62 &89.31 & 81.46 \\
\midrule
\multirow{6}{*}{DSC$\uparrow$}&\multirow{3}{*}{ LoRKD-Nano} & 4 &62.12M&80.00 &73.69 &74.96 &84.79 &89.33 &70.39 &81.11 &77.82 &79.01  \\
& & 8 &67.01M&80.06&73.80 &75.15 & 83.69& 89.28& 70.47& 81.86& 82.34& 79.58\\
& & 16 &76.78M &80.17 &73.77 &75.14 &83.49 &89.11 &70.54 &81.40 & 80.37&79.25 \\
\cmidrule{2-13}
& \multirow{3}{*}{ LoRKD-Pro}& 4 &119.15M &80.78 &75.83 &78.70 &86.93 &91.59 &74.56 &87.79 &81.80 &82.25 \\
& & 8 &129.10M& 80.56&75.79 &78.61 & 88.56& 91.75& 73.68& 87.00& 86.69& 82.83\\
& & 16 &149.02M &80.58 &75.75 &78.75 & 85.86&92.31 &73.57 &87.84 &88.10 &82.85 \\

\bottomrule[1.5pt] 
\end{tabular}}}
\end{table*}


\begin{table*}[t]\footnotesize
\centering
\caption{Ablation on $\beta$ on the pre-training segmentation dataset.
}
\label{tab:ablation-beta}
\vspace{-0.2cm}
\resizebox{0.95\textwidth}{!}{
\setlength{\tabcolsep}{3mm}{
\begin{tabular}{c|c|c|cccccccc|c}
\toprule[1.5pt]
   Metric& Model & $\beta$ & Abdomen & Brain & H\&N & LL & Pelvis & Spine & Thorax & UL& Avg \\
\midrule
\multirow{8}{*}{NSD$\uparrow$}& \multirow{4}{*}{ LoRKD-Nano}& 0.05 &67.83 &71.40 &79.48 & 80.17& 84.59&67.30 & 77.55&81.54 &76.23 \\
& & 0.1 & 68.02& 71.50& 79.64& 80.07& 84.72& 67.01& 77.75& 83.34& 76.51 \\
& & 1 & 67.97& 71.52& 79.62& 79.21& 84.83& 67.37& 77.86& 81.52& 76.24 \\
& & 10 &67.57 &70.86 &78.87 &76.93 &83.42 &65.96 &77.50 &82.21 &75.42 \\
\cmidrule{2-12}
& \multirow{4}{*}{ LoRKD-Pro}& 0.05 & 70.94&75.86 &85.21 &87.17 &88.03 &72.63 & 83.57&87.49 &81.36 \\
& & 0.1 & 70.75& 75.92 & 84.95 & 88.79& 88.51& 72.14& 82.69& 87.91& 81.46\\
& & 1 & 71.13& 75.94 & 85.17 & 86.64& 88.19& 73.19& 83.54& 84.47& 81.03\\
& & 10 &70.83 &75.87 &85.05 &80.64 &88.35 &72.54 &83.13 &88.63 & 80.63 \\
\midrule
\multirow{8}{*}{DSC$\uparrow$}&\multirow{4}{*}{ LoRKD-Nano} & 0.05 &79.93 &73.72 &74.98 &84.56 &89.25 &70.89 &81.68 &80.53 &79.44  \\
& & 0.1 &80.06&73.80 &75.15 & 83.69& 89.28& 70.47& 81.86& 82.34& 79.58\\
& & 1 &79.98 &73.84 &75.06 &82.62 &89.42 &70.79 &81.95 & 80.50&79.27 \\
& & 10 &79.90 &73.44 &74.51 &80.33 &88.17 &69.67 &81.83 & 81.39&78.65 \\
\cmidrule{2-12}
& \multirow{4}{*}{ LoRKD-Pro}& 0.05 &80.53 &75.74 &78.78 &87.56 &91.24 &74.10 &87.76 &86.27 &82.75 \\
& & 0.1 & 80.56&75.79 &78.61 & 88.56& 91.75& 73.68& 87.00& 86.69& 82.83\\
& & 1 &80.77 &75.81 &78.76 & 86.81&92.42 &74.67 &87.78 &83.29 &82.41 \\
& & 10 &80.57 &75.77 &78.68 & 80.91&91.51 &74.03 &87.40 &87.41 &82.03 \\

\bottomrule[1.5pt] 
\end{tabular}}}
\end{table*}

\begin{table}[t]
    \centering
    \caption{Ablation on rank r on classification datasets.}
    \vspace{-0.2cm}
    \resizebox{0.48\textwidth}{!}{
    \begin{tabular}{c|c|ccccc|c}
    \toprule[1.5pt]
        Rank & Pre-train & COVID & BTC & AD & Mura\_s & AUITD  & Avg\\
    \midrule
      4 & 55.08  & 85.71  & 79.95 &  75.61 & 77.98  &98.05  &83.46\\
      8 & 56.26  & 85.71  & 81.47 &  75.92 & 78.51  &98.33  &84.93\\
      16 & 56.19  & 86.21  & 82.49 &  78.81 & 78.51  &98.33  &84.87\\
    \bottomrule[1.5pt]
    \end{tabular}}
    \label{tab:rank_proformance}
    \vspace{-0.1cm}
\end{table}

\subsubsection{Knowledge Disentanglement}
\label{disentanglement}

\par{\noindent \bf Enhanced disentanglement.} 
To verify whether our method can indeed achieve knowledge decoupling between different tasks, we measure the mutual information gap (MIG) scores~\cite{chen2018isolating} across different methods. MIG is a widely used metric for assessing disentanglement. 
The results are illustrated in Figure \ref{fig:mig}, where higher MIG scores indicate a higher level of disentanglement. 
It can be observed that our method exhibits a higher level of disentanglement compared to the previous KF and other baselines.
This improvement can likely be attributed to the explicit gradient separation incorporated in our method, which effectively minimizes the interference between gradients from different tasks, thereby enhancing the specialization of the expert modules.

Additionally, we find that MTL exhibits a lower degree of disentanglement compared to STL. 
This suggests that the shared encoder architecture commonly used in MTL inadvertently leads to the entanglement of gradients from these different tasks. 
As a result, this gradient entanglement manifests as knowledge entanglement, potentially diminishing the model’s overall effectiveness in handling individual tasks.
Furthermore, STL-KD exhibits lower disentanglement compared to STL, which can be attributed to the transfer of common knowledge from the foundation model.

\par{\noindent \bf Lower Feature Similarity.}
Figure \ref{fig:cka} shows the Centered Kernel Alignment (CKA) feature similarity matrices~\cite{kornblith2019similarity} of our method and MTL on the Radimagenet dataset. 
The CKA similarity metric is a powerful tool for assessing how closely the feature representations of different tasks align with one another.
It is evident that our method exhibits significantly lower CKA feature similarity between different tasks compared to the MTL approach, which confirms the knowledge disentanglement ability of LoRKD. 
This phenomenon can be attributed to our low-rank expert modules being embedded at the convolutional level, which facilitates the simultaneous decomposition of shallow knowledge and deep knowledge.

\subsubsection{Ablation Study}
\label{ablation}

\begin{figure*}[t]
    \centering
    \subfigure{
        \centering
        \includegraphics[width=0.48\textwidth]{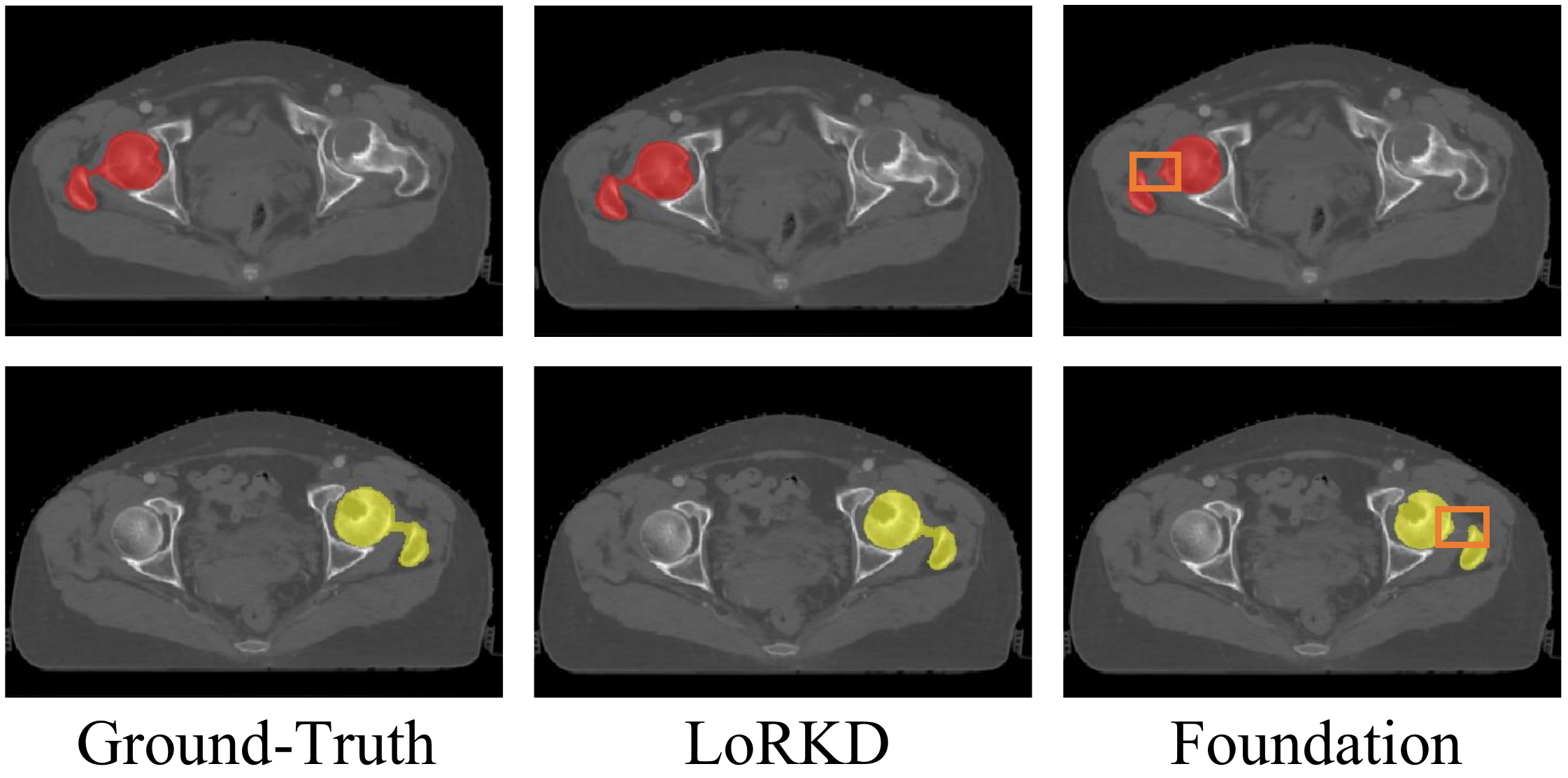}
        
        \label{fig:vis-seg-1}
    }
    \subfigure{
        \centering
        \includegraphics[width=0.48\textwidth]{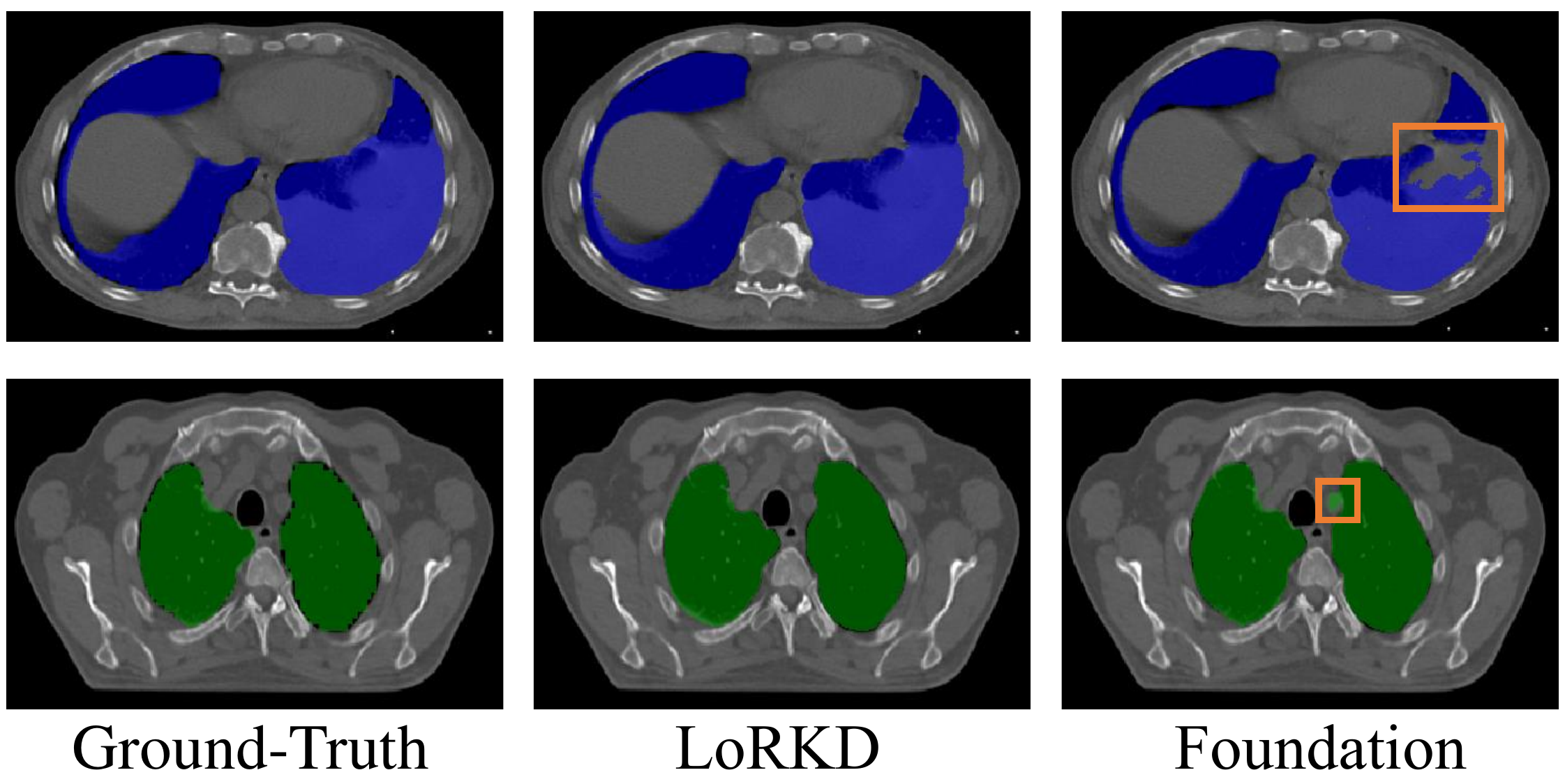}
        \label{fig:vis-seg-2}
    }
    \vspace{-0.3cm}
    \caption{Comparison of segmentation results between the decomposed model and foundation model on the SAT-DS dataset.
    Different colors represent different segmentation targets. 
    The flaws of the foundation model are highlighted in orange.
    }
    \vspace{-0.2cm}
    \label{fig:vis-seg}
\end{figure*}

\begin{figure*}[t]
    \centering
    \includegraphics[width=0.98\linewidth]{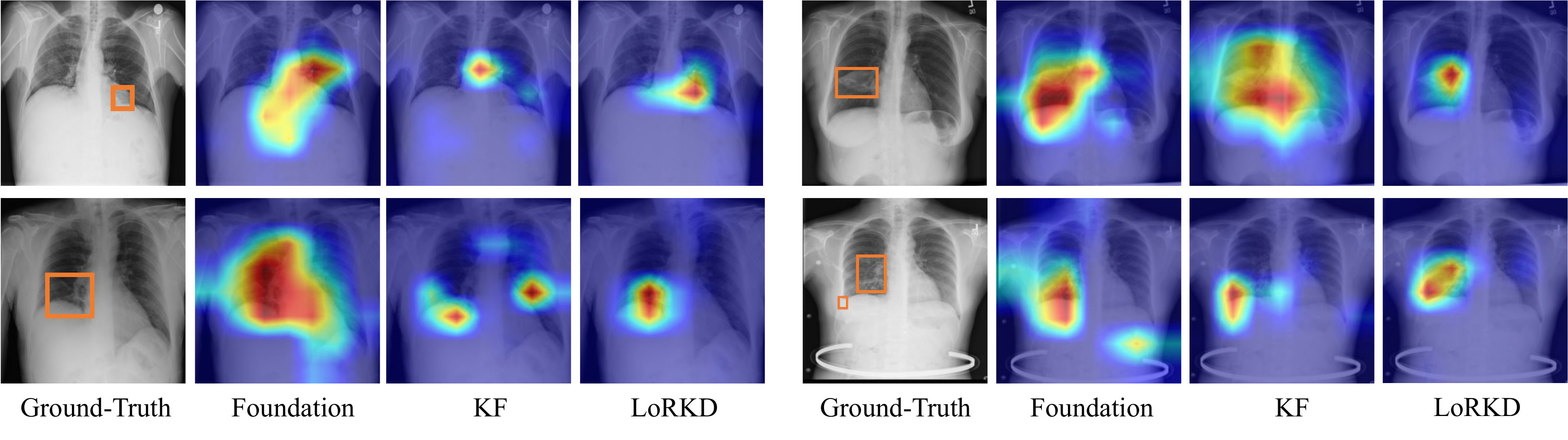}
    \vspace{-0.3cm}
    \caption{Comparison of Grad-CAM visualizations between the decomposed model and the foundation model on DET10. }
    \label{fig:grad_cam}
    \vspace{-0.43cm}
\end{figure*}

\par{\noindent \bf The impact of Rank $r$.}
The rank $r$ of low-rank experts significantly affects their representation ability and the number of parameters. 
Therefore, we conducted an ablation experiment to investigate the impact of varying the rank of low-rank experts.
The results of the segmentation task and classification task are presented in Table~\ref{tab:ablation-r} and Table~\ref{tab:rank_proformance} respectively.
For the segmentation task, whether decomposing SAT-Pro or SAT-Nano, increasing $r$ from $4$ to $8$ leads to a significant performance improvement on the pre-training dataset.
However, increasing $r$ from 8 to 16 does not yield further enhancement; in fact, the performance tends to plateau or even slightly degrade.
The results of the classification task further corroborate this conclusion, where performance generally improves from 
$r=4$ to $r=8$ but shows diminishing returns or even slight decreases when $r$ is increased to 16.
This suggests that selecting a larger 
$r$ is not necessarily better.
An appropriate rank value enables the low-rank expert module to learn distinct representations from the backbone while maintaining a manageable number of parameters.
Therefore, we selected $8$ as the base rank value.
Moreover, we again observed that the improvement in the upstream dataset is positively correlated with the improvement in transferability.

\par{\noindent \bf The impact of $\beta$.}
Table~\ref{tab:ablation-beta} shows the ablation experiment about the impact of the trade-off parameter $\beta$ between the $\mathcal{L}_{task}$ and $\mathcal{L}_{transfer}$.
We observe that increasing $\beta$ from 0.05 to 1 does not lead to significant performance fluctuations, but further increasing $\beta$ to 10 results in a noticeable performance drop.
This indicates that maintaining an appropriate $\beta$ value is critical for optimizing decomposition performance, as an excessively large value can negatively impact the training process.
Overall, a $\beta$ value of 0.1 achieves an appropriate balance between the two loss functions, consistently yielding the best results.

\subsubsection{Visualization}
\label{vis}

\par{\noindent \bf Stronger Specialization.}
In this subsection, we visualize the experimental results and analyze the specialization brought by knowledge decomposition.
Figure~\ref{fig:vis-seg} presents segmentation results, comparing the ground truth with images segmented by our LoRKD model and the foundation model (SAT-Pro). 
Different colors represent distinct segmentation targets: the left side is the segmentation of ``head of femur", red and yellow denote the right and left femur, respectively; the right side is the segmentation of ``thoracic cavity", blue and green correspond to different slices respectively. 
The foundation model exhibits several noticeable segmentation flaws, including clearly missing parts of the target regions and over-segmenting certain areas.
In contrast, our \shortname demonstrates stronger specialization, producing segmentation results that closely align with the ground truth.

Taking the DET10 dataset as an example, we evaluate the differences in the activated regions between the decomposed expert model and the foundation model during the prediction process from the perspective of Grad-CAM~\cite{selvaraju2017grad}. 
Grad-CAM highlights the regions of an input image most relevant to a neural network's decision, offering insights into how the model interprets the image.
As illustrated in Figure~\ref{fig:grad_cam}, the visualization results reveal notable differences between different models. 
The foundation model tends to focus on broader, less specific regions of the image. 
This broad focus is indicative of the model's ability to capture general features across a wide range of tasks, yet it lacks the precision required for more specialized applications. 
The KF model's focus is more refined than the foundation model but remains less precise than our decomposed expert model.
In contrast, our decomposed expert model exhibits a more refined focus, concentrating on smaller, more precise regions that are highly relevant to the specific task at hand. 
This precise localization indicates a higher degree of specialization.
These findings underscore the effectiveness of our approach in improving specialization and efficiency, particularly in scenarios where precise region identification is crucial.

\section{Conclusion}

In this paper, we propose a new perspective called knowledge decomposition, aimed at reducing the deployment costs and enhancing specialization for medical foundation models. 
We develop low-rank expert modules and efficient gradient separation convolution to decompose the foundation model into multiple lightweight expert models. 
Our method includes two variants: LoRKD-balance and LoRKD-imbalance.
The former assigns a low-rank expert module of the same rank to each task, while the latter adaptively adjusts the rank of each module based on task complexity.
The decomposition performance on upstream tasks and the transfer performance on downstream tasks fully demonstrate that \shortname can effectively alleviate the conflict of heterogeneous data, achieving cost reduction and performance improvement simultaneously.
We hope this research offers valuable insights for advancing the development and deployment of medical foundation models.

\ifCLASSOPTIONcompsoc
  \section*{Acknowledgments}
\else
  \section*{Acknowledgment}
\fi


\bibliographystyle{ieeetr}
\bibliography{ref}

\vskip -2\baselineskip plus -1fil
\begin{IEEEbiography}[{\includegraphics[width=1in,height=1.25in,clip,keepaspectratio]{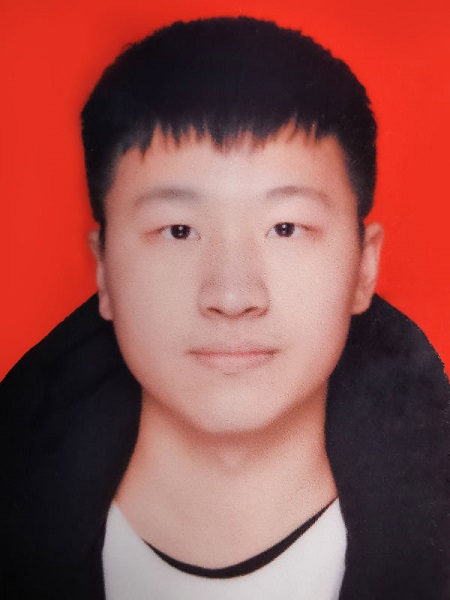}}]{Haolin Li}
	received a B.S. degree from University of Electronic Science and Technology of China, in 2023. 
    He is currently working toward the PhD degree from Fudan University, advised by Prof. J. Yao and Prof. Y. Zhang.  
    His research interests include computer vision and AI for Healthcare.
\end{IEEEbiography}
\vskip -2\baselineskip plus -1fil
\begin{IEEEbiography}[{\includegraphics[width=1in,height=1.25in,clip,keepaspectratio]{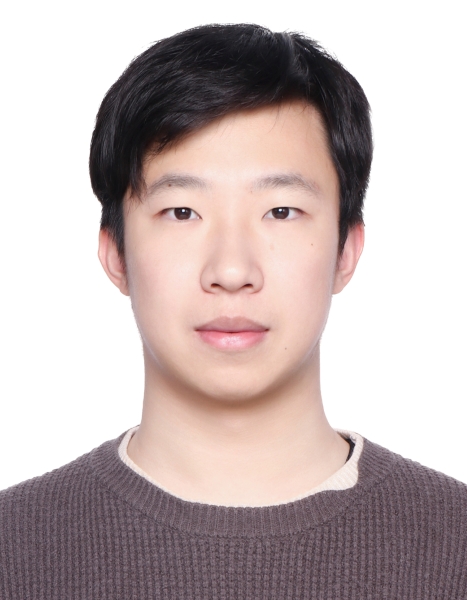}}]{Yuhang Zhou}
	received a B.S. degree from University of Electronic Science and Technology of China, in 2019. 
    He is currently working toward the PhD degree from Shanghai Jiao Tong University, advised by Prof. J. Yao and Prof. Y. Zhang.  
    His research interests include computer vision, machine learning and AI for Healthcare.
\end{IEEEbiography}
\vskip -2\baselineskip plus -1fil
\begin{IEEEbiography}[{\includegraphics[width=1in,height=1.25in,clip,keepaspectratio]{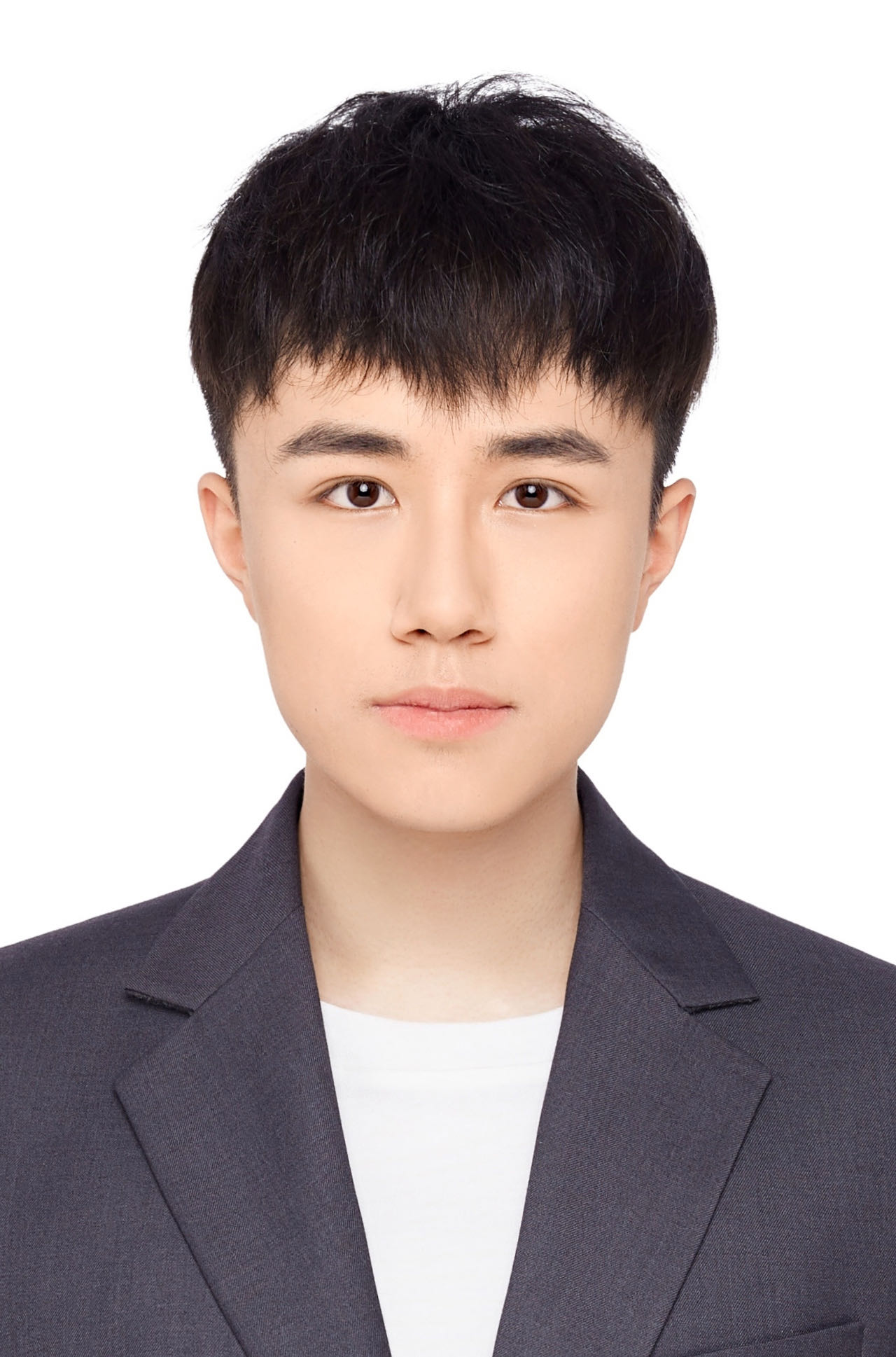}}]{Ziheng Zhao} 
    received a B.S. degree from Shanghai Jiao Tong University, in 2021. 
    He is currently working toward the PhD degree from Shanghai Jiao Tong University, advised by Prof. W. Xie and Prof. Y. Zhang.  
    His research interests include computer vision and AI for Healthcare.
\end{IEEEbiography}
\vskip -2\baselineskip plus -1fil
\begin{IEEEbiography}[{\includegraphics[width=1in,height=1.25in,clip,keepaspectratio]{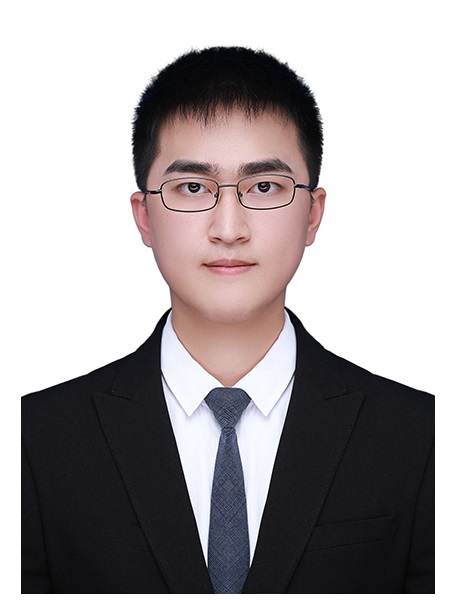}}]{Siyuan Du} 
    received a B.S. degree from University of Electronic Science and Technology of China, in 2023. 
    He is currently working toward the PhD degree from Fudan University, advised by Prof. J. Yao and Prof. Y. Zhang.  
    His research interests include computer vision and AI for Healthcare.
\end{IEEEbiography}
\vskip -2\baselineskip plus -1fil
\begin{IEEEbiography}[{\includegraphics[width=1in,height=1.25in,clip,keepaspectratio]{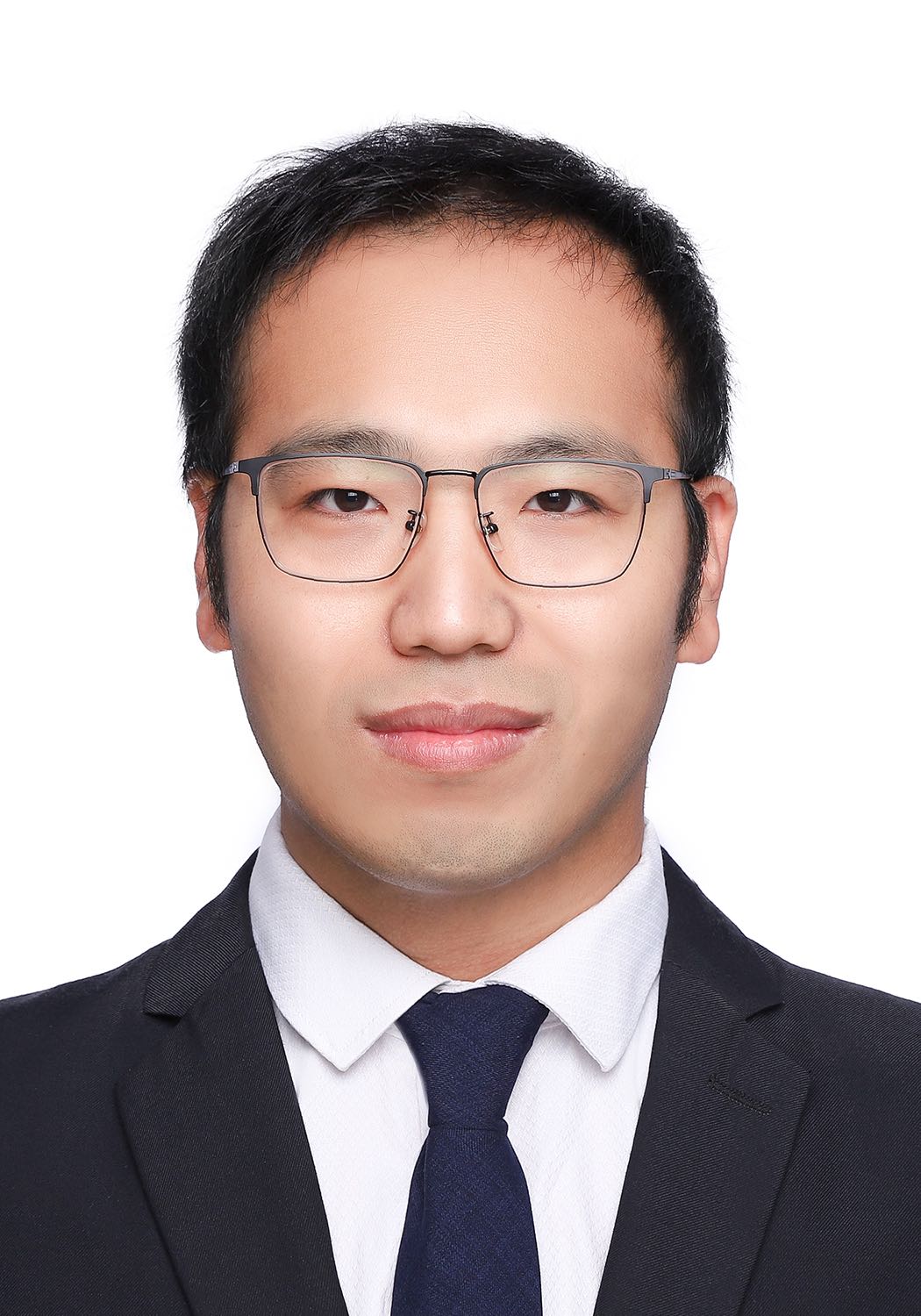}}]{Jiangchao Yao} is an Assistant Professor of Shanghai Jiao Tong University, Shnaghai China. 
He received the B.S. degree in information engineering from South China University of Technology, Guangzhou, China, in 2013. He got a dual Ph.D. degree under the supervision of Ya Zhang in Shanghai Jiao Tong University and Ivor W. Tsang in University of Technology Sydney.
His research interests include deep representation learning and robust machine learning.
\end{IEEEbiography}
\vskip -2\baselineskip plus -1fil
\begin{IEEEbiography}[{\includegraphics[width=1in,height=1.25in,clip,keepaspectratio]{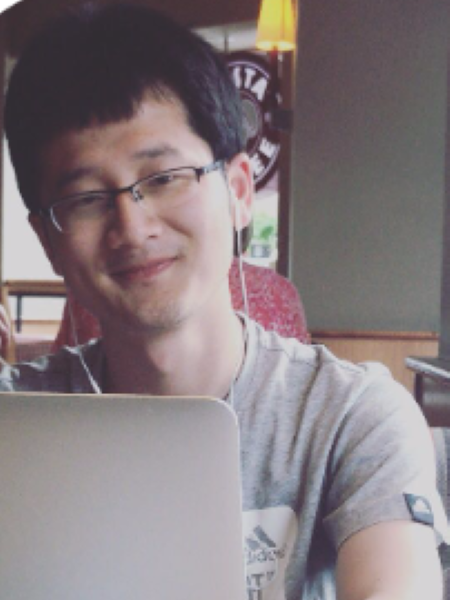}}]{Weidi Xie} is an Associate Professor of Shanghai Jiao Tong University, Shnaghai China. 
Prior to that, He completed D.Phil at Visual Geometry Group, University of Oxford, advised by Professor Andrew Zisserman (VGG), and Professor Alison Noble (BioMedIA).
His research interests include computer vision, deep learning, and biomedical image analysis.
\end{IEEEbiography}
\vskip -2\baselineskip plus -1fil
\begin{IEEEbiography}[{\includegraphics[width=1in,height=1.25in,clip,keepaspectratio]{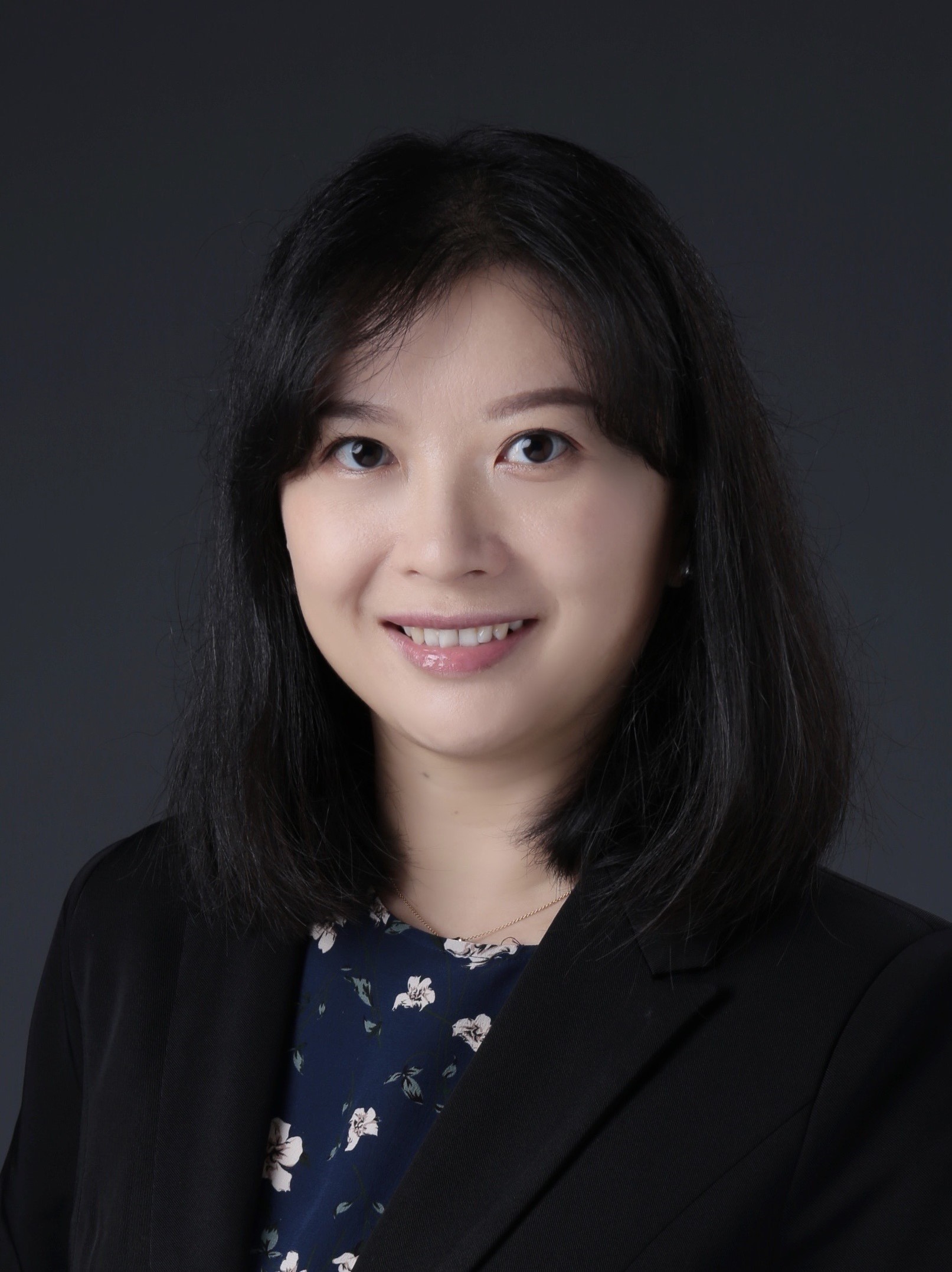}}]{Ya Zhang} (Member, IEEE) received the B.S. degree from Tsinghua University and the Ph.D. degree in information sciences and technology from the Pennsylvania State University. Since March 2010, she has been a professor with Cooperative Medianet Innovation Center, Shanghai Jiao Tong University. Prior to that, she worked with Lawrence Berkeley National Laboratory, University of Kansas, and Yahoo! Labs. Her research interest is mainly on data mining and machine learning, with applications to information retrieval, web mining, and multimedia analysis.
\end{IEEEbiography}
\vskip -2\baselineskip plus -1fil
\begin{IEEEbiography}[{\includegraphics[width=1in,height=1.25in,clip,keepaspectratio]{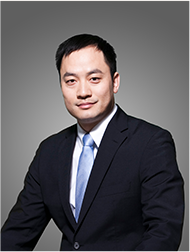}}]{Yanfeng Wang} received the B.E. degree in information engineering from the University of PLA, Beijing, China, and the M.S. and Ph.D. degrees in business management from the Antai College of Economics and Management, Shanghai Jiao Tong University, Shanghai, China. He is currently the Vice Director of the Cooperative Medianet Innovation Center and also the Vice Dean of the School of Electrical and Information Engineering, Shanghai Jiao Tong University. His research interests mainly include media big data and emerging commercial applications of information technology.
\end{IEEEbiography}


\end{document}


\title{LoRKD: Low-Rank Knowledge Decomposition for Medical Foundation Models}

\author{Haolin~Li,
        Yuhang~Zhou,
        Ziheng~Zhao,
        Siyuan~Du,
        Jiangchao~Yao,
        Weidi~Xie,
        Ya~Zhang,
        and Yanfeng~Wang
\IEEEcompsocitemizethanks{
\vspace{-5pt}
\IEEEcompsocthanksitem 
Haolin Li and Siyuan Du are with the School of Computer Science, Fudan University and Shanghai
AI Laboratory, Shanghai 200437, China (e-mail: {23110240025, 23110240011}@m.fudan.edu.cn).
\IEEEcompsocthanksitem Yuhang Zhou, Ziheng Zhao, and Jiangchao Yao are with the Cooperative
Medianet Innovation Center, Shanghai Jiao Tong University and Shanghai
AI Laboratory, Shanghai 200240, China. (E-mail: {zhouyuhang, Zhao\_Ziheng, sunarker}@sjtu.edu.cn).
\IEEEcompsocthanksitem Weidi Xie, Ya Zhang and Yanfeng Wang are with the AI School, Shanghai Jiao Tong University and Shanghai
AI Laboratory, Shanghai 200240, China. (E-mail: {weidi, ya\_zhang, wangyanfeng}@sjtu.edu.cn).
\IEEEcompsocthanksitem Jiangchao Yao and Yanfeng Wang are the corresponding authors.
}
}

\markboth{Submitted to IEEE TPAMI}%
{Shell \MakeLowercase{\textit{et al.}}: Bare Advanced Demo of IEEEtran.cls for IEEE Computer Society Journals}

\maketitle


%
\IEEEpeerreviewmaketitle

\section{Pre-training datasets}
\subsection{Segmentation Pre-training datasets}
\begin{figure*}[h]
\centering
\begin{minipage}{0.45\linewidth}
\centerline{\includegraphics[width=0.98\linewidth]{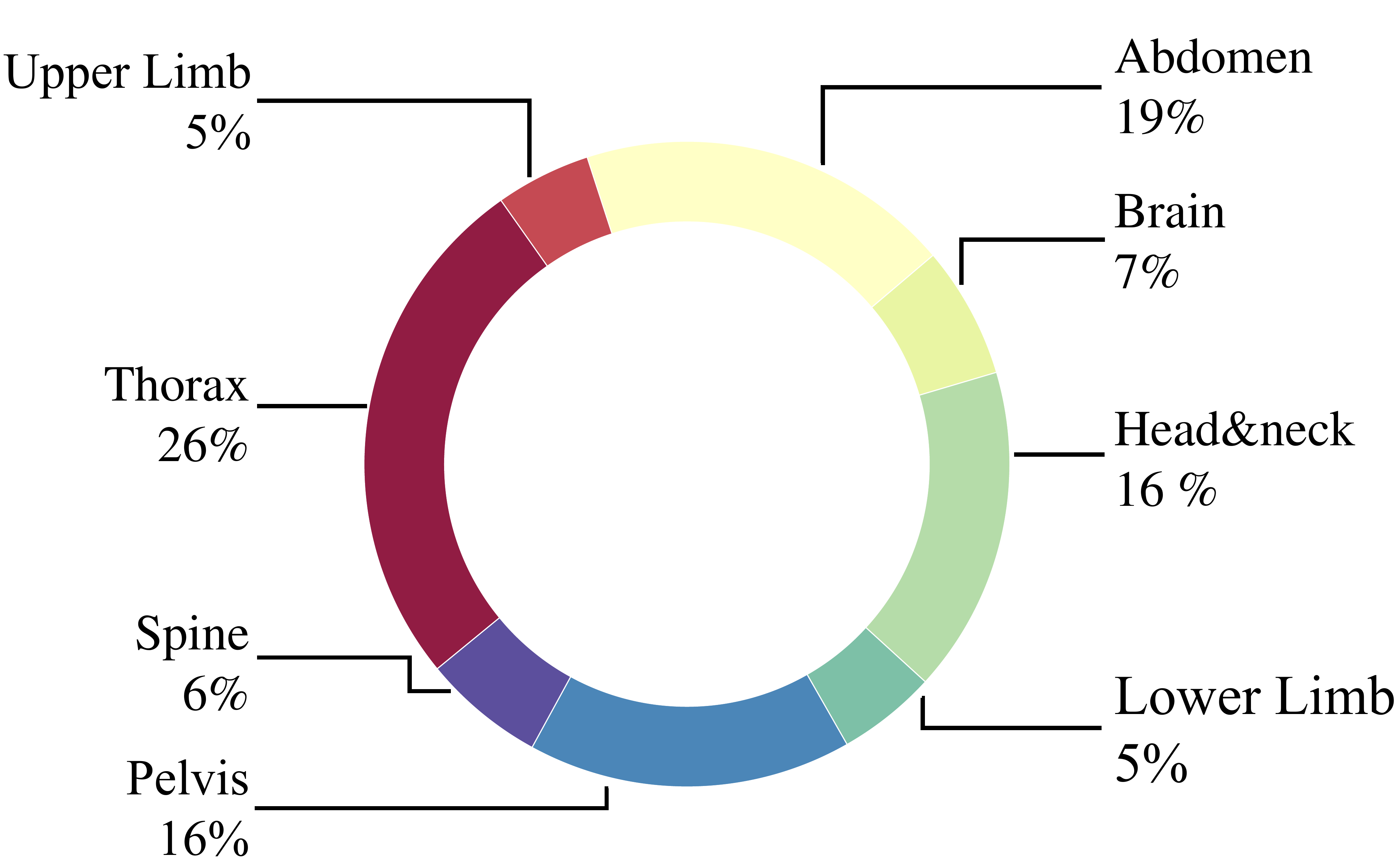}}
\end{minipage}
\hfill
\begin{minipage}{.48\linewidth}
\resizebox{0.98\textwidth}{!}{
\begin{tabular}{cccccc}
\multicolumn{6}{c}{\textbf{SAT-DS}~\cite{zhao2023one}}  \\
\toprule[1.5pt]
Task ID & Region & Modality & Labels & Number & Masks  \\ 
\midrule 
1  & Abdomen  & CT\&MRI & 49 & 5128 & 38583\\ 
2  & Brain  & CT\&MRI & 131 & 1804 & 6884\\ 
3  & Head\&neck  & CT\&MRI & 74 & 4466 & 15043   \\ 
4  & Lower Limb  & CT\&MRI  & 10  & 1352 & 4452   \\ 
5  & Pelvis  & CT\&MRI  & 23 & 4423 & 25241 \\ 
6  & Spine &  CT\&MRI & 54 &  1667 & 40226\\ 
7  & Thorax& CT\&MRI  & 79 &7135 & 74771 \\ 
8  & Upper Limb &CT  & 8 & 1301 & 9616 \\ 
\bottomrule[1.5pt]
\end{tabular}}
\end{minipage}
\caption{Data distribution in SAT-DS.
Images including labels from multiple regions are calculated multiple times in the ``Number'' column.
}
\label{sat}
\end{figure*}

The dataset SAT-DS~\cite{zhao2023one} consists of 13303 3D images, 428 labels, and 214816 segmentation masks, covering 8 tasks (regions) and 2 common modalities. 
The distribution of data is shown in Figure~\ref{sat}. 
In our experiments, we decompose the pre-trained models on SAT-DS into eight lightweight expert models corresponding to task IDs.
Detailed information about the 49 datasets that make up SAT-DS, including the number of scans, classes, annotations, and associated body regions, is provided in Table~\ref{tab:dataset_details}. 
For more comprehensive details, please refer to the original paper.

For the ``specialist'' baseline nnUNet, we train 49 independent nnUNet models on the 49 datasets that comprise SAT-DS. 
Since the method automatically configures the model based on the dataset, different datasets will have various configurations.
The specific configuration of each model is shown in Table~\ref{tab:nnunet_config}.

\subsection{Classification Pre-training datasets}

\begin{figure*}[h]
\begin{minipage}{0.45\linewidth}
 \centerline{\includegraphics[width=0.98\linewidth]{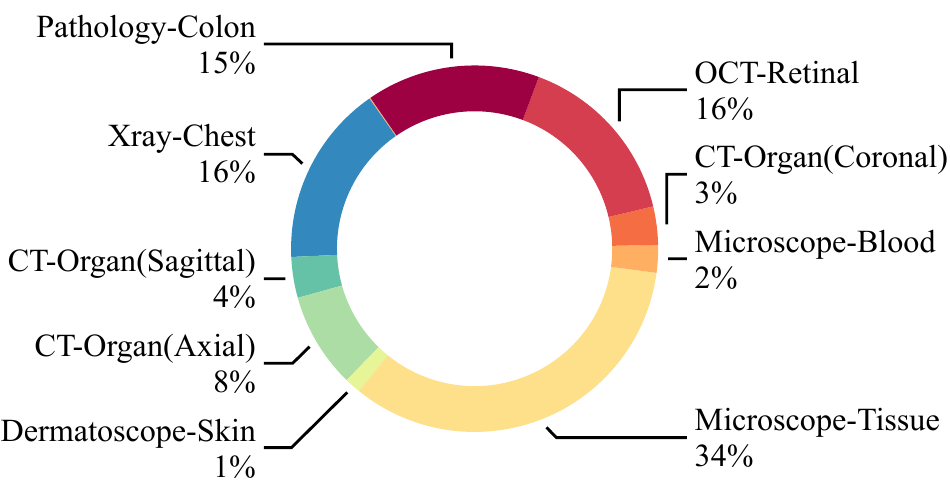}}
\end{minipage}
\hfill
\begin{minipage}{.48\linewidth}
\resizebox{0.98\textwidth}{!}{
\begin{tabular}{cccccc}
\multicolumn{6}{c}{\textbf{MedMnist}}  \\
\toprule[1.5pt]
Task ID & Name & Modality & Region & Labels & Number  \\ 
\midrule
1  & Colon  & Pathology   & Colon  & 9  & 107180  \\ 
2  & Retinal  & OCT & Eye & 4 & 109309  \\ 
3  & OrganC  & CT & Abdomen & 11 & 23660   \\ 
4  & Cell  & Microscope  &  Blood  & 8  & 17092   \\ 
5  & Breast  & Ultrasound  &  Breast  & 2 & 780 \\ 
6  & Tissue & Microscope  &  Kidney cortex  & 8 & 236386 \\ 
7  & Skin & Dermatoscope  &  Skin  & 7 & 10015 \\ 
8  & OrganA  & CT  &  Abdomen  & 11 & 58850 \\ 
9  & OrganS & CT  &  Abdomen  & 11 & 25221 \\
10  & Chest & Xray  &  Chest  & 2 &  112120 \\ 
\bottomrule[1.5pt]
\end{tabular}}
\end{minipage}
\caption{Data distribution in MedMnist.}
\label{medmnist}
\end{figure*}

\begin{figure*}[h]
\begin{minipage}{0.45\linewidth}
 \centerline{\includegraphics[width=0.98\linewidth]{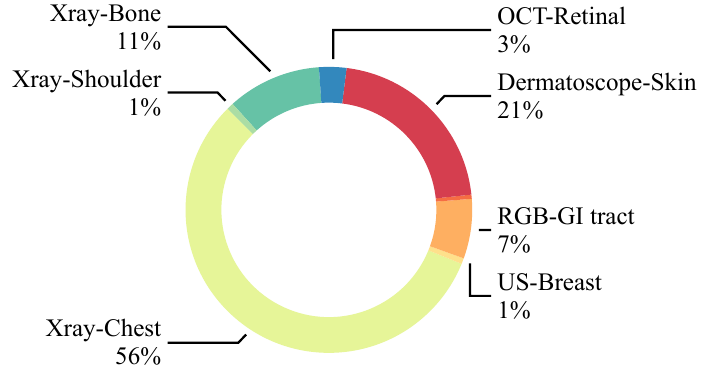}}
\end{minipage}
\hfill
\begin{minipage}{.48\linewidth}
\resizebox{0.98\textwidth}{!}{
\begin{tabular}{cccccc}
\multicolumn{6}{c}{\textbf{Med-MT}}  \\
\toprule[1.5pt]
Task ID & Name & Modality & Region & Labels & Number  \\ 
\midrule
1  & Retinal~\cite{Aptos}  & OCT   & Eye  & 5  & 3662  \\ 
2  & Skin~\cite{ISIC_2019}  & Dermatoscope & Skin & 3 & 25331  \\ 
3  & Breast~\cite{al2020busi}  & Ultrasound & Breast & 8 & 780   \\ 
4  & GI tract~\cite{Kvasir_v2}  & RGB  &  Gastrointestinal  & 8  & 8000   \\ 
5  & Lung~\cite{shenzhen_x-ray}  & Xray  &  Chest  & 2 & 566 \\ 
6  & Shoulder~\cite{Shouder_x-ray} &  Xray &   Shoulder & 4 & 945 \\ 
7  & Lung~\cite{nguyen2022vindr} &  Xray &   Chest & 15 &  67914 \\ 
8  & Bone~\cite{halabi2019rsna} &  Xray &  Bone  & 12 & 12611 \\ 
\bottomrule[1.5pt]
\end{tabular}}
\end{minipage}
\caption{Data distribution in Med-MT.}
\label{medmt}
\end{figure*}

\begin{itemize}
    \item The dataset RadImageNet~\cite{mei2022radimagenet} consists of 1.35 million images, covering 11 tasks and 3 common modalities. The distribution of diseases is shown in Figure~\ref{rad}. In our experiments, we decompose the pre-trained models on RadImageNet~\cite{mei2022radimagenet} (which have been publicly released by the authors) into 11 lightweight expert models corresponding to task IDs.
    
    \item The dataset MedMnist is selected from MedMnistV2~\cite{yang2023medmnist}, consisting of 705,689 images, covering 10 tasks and 7 different modalities. The distribution of diseases is shown in Figure~\ref{medmnist}. In our experiments, we decompose the fully pre-trained models on MedMnist into 10 lightweight expert models corresponding to task IDs.
    
    \item The dataset Med-ML is a multi-task dataset we constructed, consisting of 119,655 images, covering 8 tasks and 5 different modalities, including APTOS~\cite{Aptos}, ISIC~\cite{ISIC_2019}, BUSI~\cite{al2020busi}, Kvasir~\cite{Kvasir_v2}, Shenzhen X-ray~\cite{shenzhen_x-ray}, Shoulder X-ray~\cite{Shouder_x-ray}, VinDr~\cite{nguyen2022vindr} and Bone~\cite{halabi2019rsna}. The distribution of diseases is shown in Figure~\ref{medmt}. In our experiments, we  decompose the fully pre-trained models on Med-ML into 8 lightweight expert models corresponding to task IDs.
\end{itemize}

\begin{figure*}[h]
\centering
\begin{minipage}{0.45\linewidth}
\centerline{\includegraphics[width=0.98\linewidth]{figures/rad2.pdf}}
\end{minipage}
\hfill
\begin{minipage}{.48\linewidth}
\resizebox{0.98\textwidth}{!}{
\begin{tabular}{cccccc}
\multicolumn{6}{c}{\textbf{Radimagenet}~\cite{mei2022radimagenet}}  \\
\toprule[1.5pt]
Task ID & Name & Modality & Region & Labels & Number  \\ 
\midrule
1  & Lung  & CT   & Chest  & 6 & 152528 \\ 
2  & Abdomen  & CT & Abdomen & 28 & 139825\\ 
3  & Thyroid  & Ultrasound & Neck & 2 & 92599   \\ 
4  & Abdomen  & Ultrasound  &  Abdomen  & 13  & 297286   \\ 
5  & Knee  & MRI  &  Knee  & 18 & 179555 \\ 
6  & Shoulder &  MRI & Shoulder   & 14 &  52407\\ 
7  &  Spine& MRI  & Spine   & 9 &71674  \\ 
8  & Ankle &MRI  & Foot   & 25 & 181603 \\ 
9  & Abdomen &  MRI & Abdomen   & 26 & 91348 \\ 
10  & Brain &MRI   &  Head  & 10 & 44671 \\ 
11  & Hip &  MRI &  Hip  & 14 &  51417\\ 
\bottomrule[1.5pt]
\end{tabular}}
\end{minipage}
\caption{Data distribution in Radimagenet.}
\label{rad}
\end{figure*}

\section{Downstream Datasets}

\subsection{Segmentation Downstream Datasets}

In the experiments, the downstream datasets we used for the segmentation task include CHAOS\_CT~\cite{kavur2021chaos}, COVID19~\cite{ma2021toward}, MSD\_Liver, MSD\_Spleen, and MSD\_Hippocampus~\cite{antonelli2022medical}. 
These datasets cover five common modalities and are used to thoroughly validate the effectiveness and generalization of our method. 
The detailed description of these datasets is presented in Figure~\ref{downstream}.

\begin{figure*}[h]
\begin{minipage}{0.45\linewidth}
 \centerline{\includegraphics[width=0.98\linewidth]{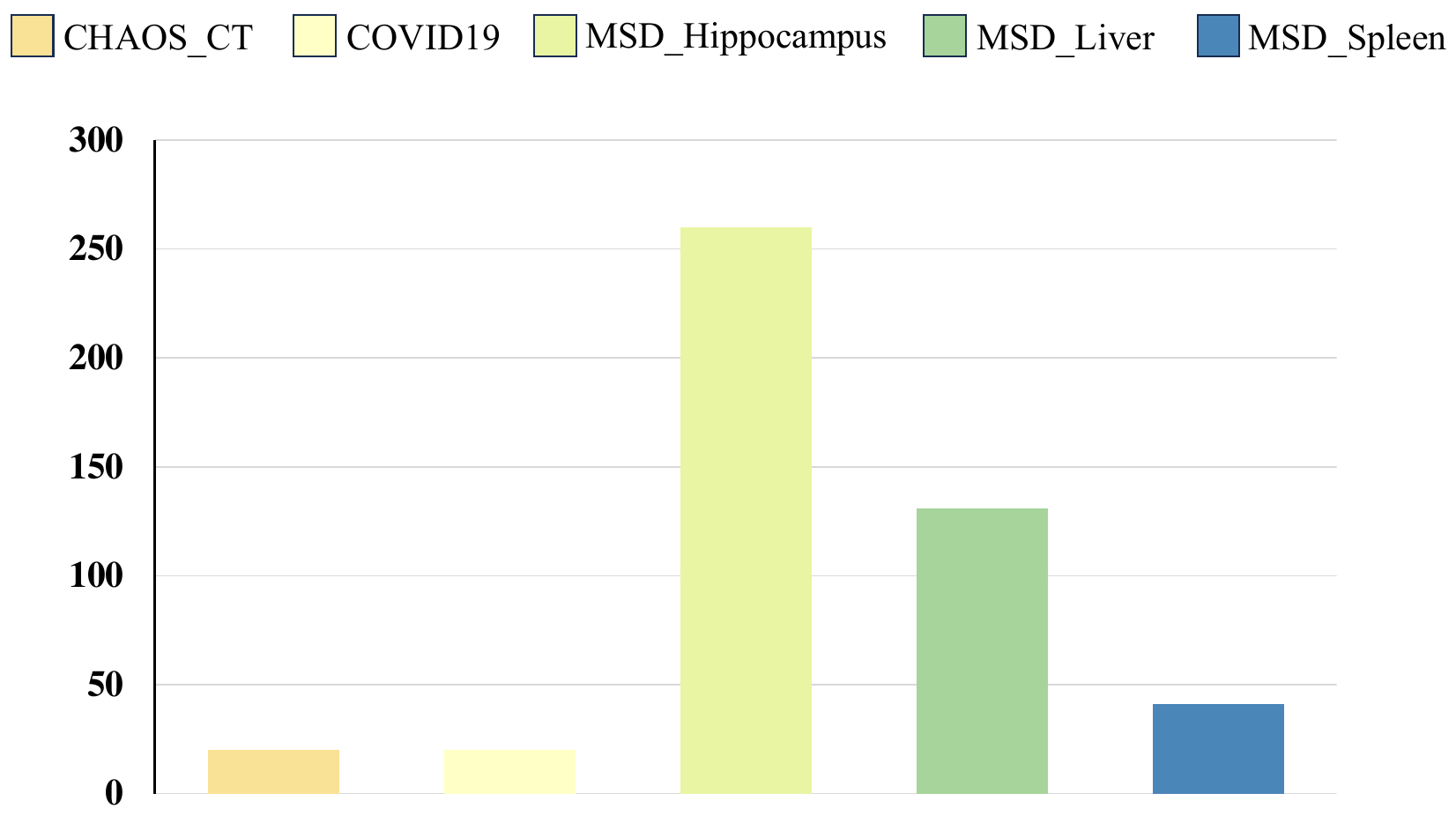}}
\end{minipage}
\hfill
\begin{minipage}{.48\linewidth}
\resizebox{0.98\textwidth}{!}{
\begin{tabular}{ccccccc}
\multicolumn{6}{c}{\textbf{Five Downstream Datasets on Segmentation}}  \\
\toprule[1.5pt]
Task ID & Name & Modality & Region & Labels & Number & Masks  \\ 
\midrule
1  & CHAOS\_CT~\cite{kavur2021chaos}  & CT   & Abdomen  & 1  & 20 & 20  \\ 
2  & COVID19~\cite{ma2021toward}  & CT & Thorax & 4 & 20 & 80  \\ 
3  & MSD\_Hippocampus~\cite{antonelli2022medical}  & MRI & Brain & 3 & 260 & 780  \\ 
4  & MSD\_Liver~\cite{antonelli2022medical}  & CT  &  Abdomen  & 2  & 131 & 262   \\ 
5  & MSD\_Spleen~\cite{antonelli2022medical}  & CT  &  Abdomen  & 1 & 41 & 41 \\ 
\bottomrule[1.5pt]
\end{tabular}}
\end{minipage}
\caption{Five downstream datasets used in our segmentation experiment.}
\label{downstream-seg}
\end{figure*}

\subsection{Classification Downstream Datasets}

\begin{figure*}[h]
\begin{minipage}{0.45\linewidth}
 \centerline{\includegraphics[width=0.98\linewidth]{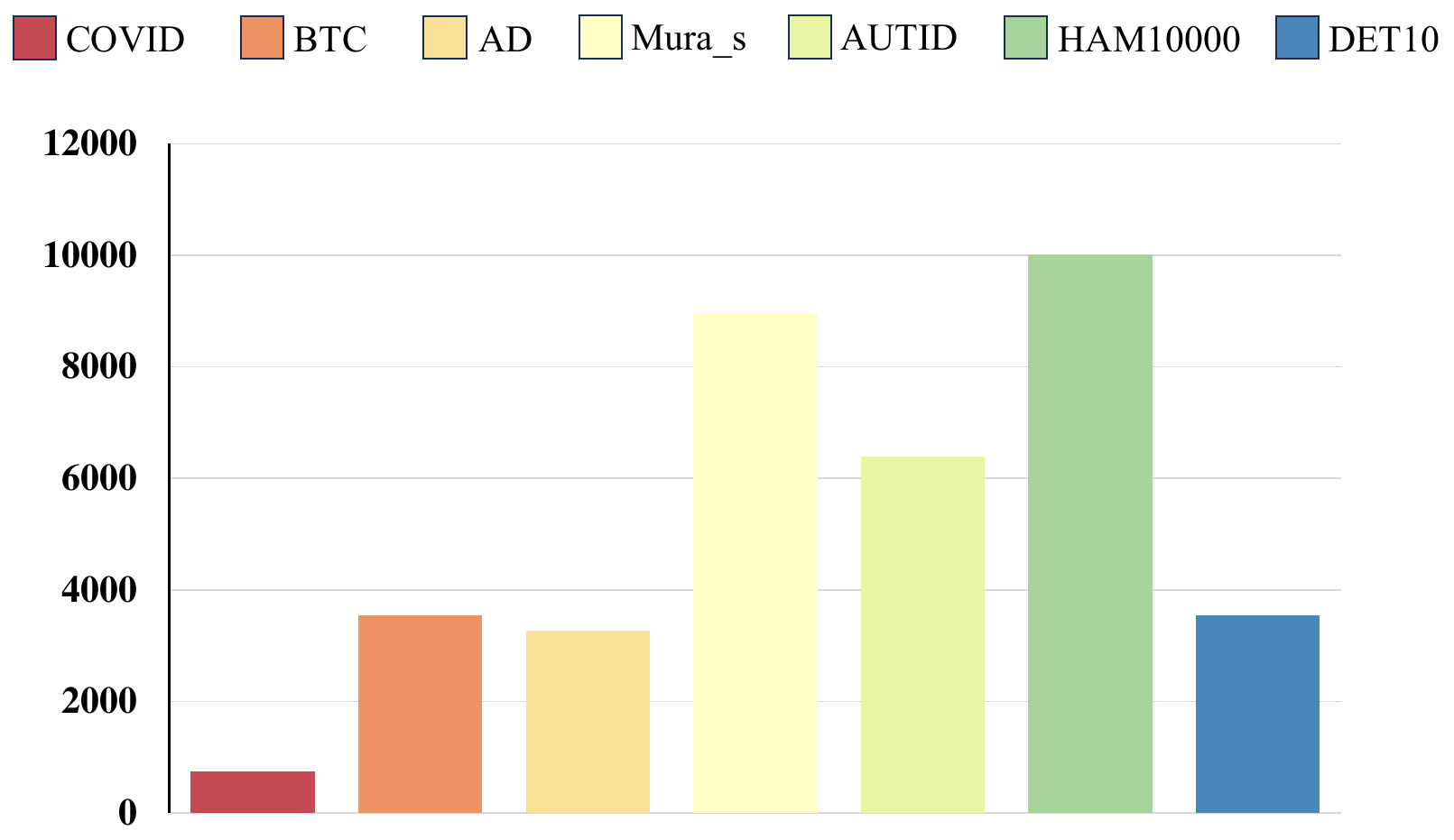}}
\end{minipage}
\hfill
\begin{minipage}{.48\linewidth}
\resizebox{0.98\textwidth}{!}{
\begin{tabular}{cccccc}
\multicolumn{6}{c}{\textbf{Seven Downstream Datasets}}  \\
\toprule[1.5pt]
Task ID & Name & Modality & Region & Labels & Number  \\ 
\midrule
1  & COVID~\cite{xingyi2020covid_CT}  & CT   & Chest  & 2  & 746  \\ 
2  & BTC~\cite{saleh2020BTC}  & MRI & Head & 4 & 3538  \\ 
3  & AD~\cite{AD}  & MRI & Head & 4 & 3264   \\ 
4  & Mura\_shoulder~\cite{rajpurkar2017mura}  & MRI  &  Shoulder  & 2  & 8942   \\ 
5  & AUTID~\cite{AUITD}  & Ultrasound  &  Neck  & 3 & 6400 \\ 
6  & HAM10000~\cite{tschandl2018ham10000} & Dermatoscope  &  Skin  & 7 &  10015\\ 
7  & DET10~\cite{liu2020chestxdet10} &  Xray & Chest   & 10 & 3543 \\ 
\bottomrule[1.5pt]
\end{tabular}}
\end{minipage}
\caption{Seven downstream datasets used in our classification experiment.}
\label{downstream}
\end{figure*}

In the experiments, the downstream datasets we used include  COVID~\cite{xingyi2020covid_CT}, BTC~\cite{saleh2020BTC}, AD~\cite{AD}, Mura~\cite{rajpurkar2017mura}, AUITD~\cite{AUITD}, HAM10000~\cite{tschandl2018ham10000}, and DET10~\cite{liu2020chestxdet10}. These datasets cover five common modalities and are used to thoroughly validate the effectiveness and generalization of our method. The description of these datasets is shown in Figure~\ref{downstream}.

\section{Detailed nnUNet baseline Configuration}
In the segmentation task, we employ the state-of-the-art specialist model nnUNet~\cite{isensee2021nnu} as a baseline.
The results on the pretraining dataset SAT-DS was the aggregated results of 49 independent nnUNet models trained on each sub-datasets.
Since nnUNet adaptively adjusts the configuration of the model according to the dataset, the configuration on each sub-dataset is different. 
We list the specific configurations in Table~\ref{tab:nnunet_config}.

\section{Correspondence between experts and datasets.}

For STL-based methods, because they train models independently for each task in the pre-training dataset, the trained models can be considered as ``expert models" lacking general knowledge. In this case, fine-tuning is only performed when the downstream task matches the model. In the main text, the ``-" symbol is used to indicate whether there is a match.

For MTL-based methods, we fine-tune their shared encoders on all downstream datasets since MTL-based methods do not generate task-specific experts.

For KF and our method LoRKD, we fine-tune the corresponding expert models on each downstream dataset.
The correspondence between expert models and downstream datasets can be seen in Table~\ref{corr}.
The symbol $^{\dagger}$ indicates the absence of a corresponding expert model (due to task or modality mismatch). Following the work of~\cite{yang2022factorizing}, in such cases, we fine-tune a shared backbone that incorporates general knowledge learned from multiple tasks.

\begin{table*}[h]
\caption{Correspondence between expert models and downstream datasets.}
\resizebox{1.0\textwidth}{!}{
\setlength{\tabcolsep}{5.8mm}{
\begin{tabular}{c|ccccccc}
\toprule[1.5pt]
Pre-trained data& COVID~\cite{xingyi2020covid_CT} & BTC~\cite{saleh2020BTC}  & AD~\cite{AD} & Mura\_s~\cite{rajpurkar2017mura} & AUTID~\cite{AUITD} & HAM10000~\cite{tschandl2018ham10000} & DET10~\cite{liu2020chestxdet10}  \\ 
\midrule
Radimagenet  & Expert\_1  &  Expert\_10 & Expert\_10  & Expert\_6 &  Expert\_3 & ${^\dagger}$ & Expert\_1\\ 
MedMnist  & Expert\_10  & ${^\dagger}$ & ${^\dagger}$  & ${^\dagger}$ & Expert\_5  & Expert\_7 & Expert\_10  \\ 
Med-MT   &  ${^\dagger}$ & ${^\dagger}$ & ${^\dagger}$  & ${^\dagger}$ & ${^\dagger}$  & Expert\_2 & ${^\dagger}$ \\ 
\bottomrule[1.5pt]
\end{tabular}}}
\label{corr}
\end{table*}

\section{Visualization of Segmentation Results}
\begin{figure*}[t]
    \centering
    \includegraphics[width=0.98\linewidth]{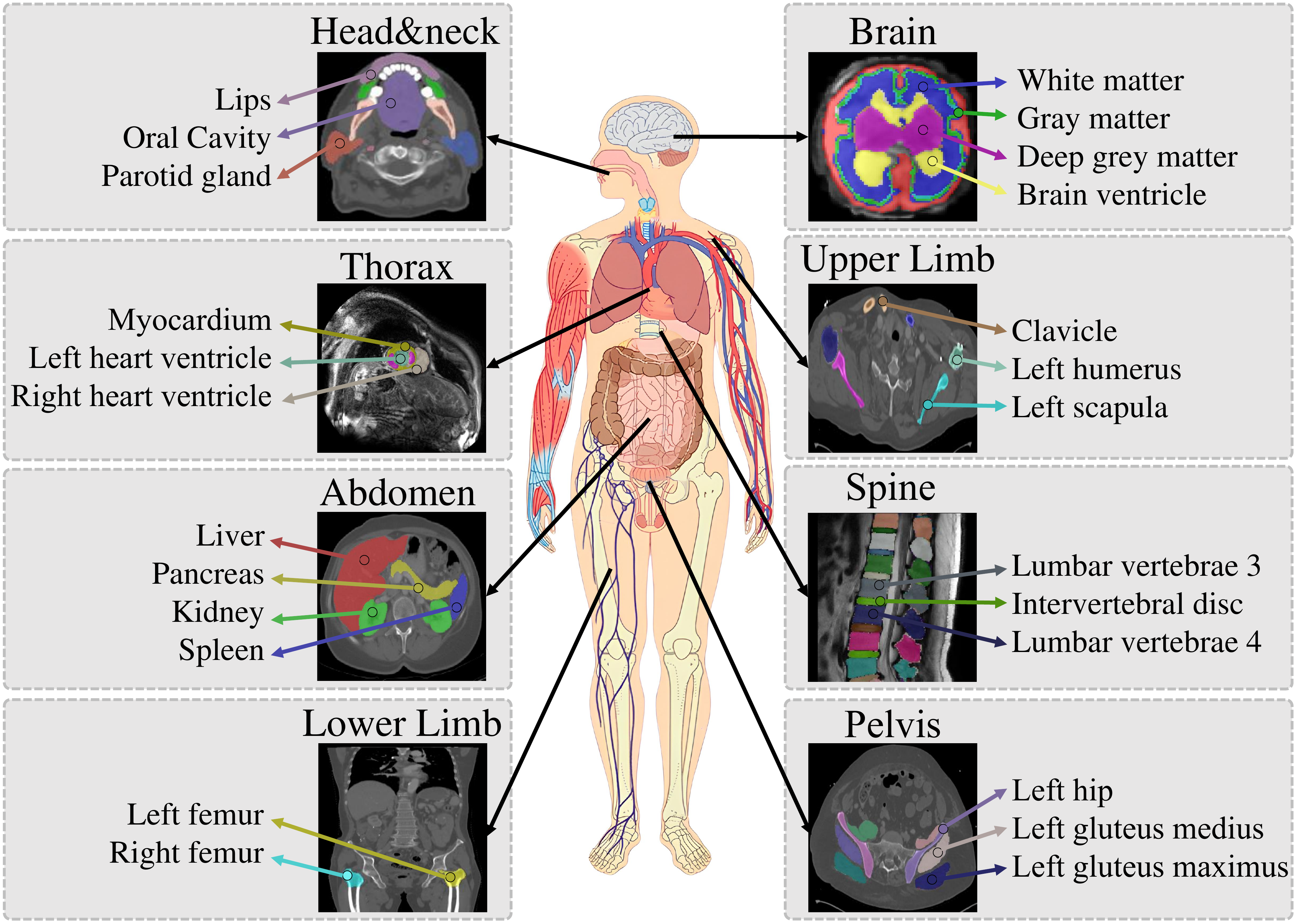}
    \vspace{-0.3cm}
    \caption{Visualization of segmentation results. 
    Segmentation targets are highlighted with different colors.
    All results are obtained using our LoRKD*-Pro model.}
    \label{fig:vis-seg}
    \vspace{-0.43cm}
\end{figure*}

In this section, we visualize the experimental results.
We present the segmentation results in Figure~\ref{fig:vis-seg}, which includes images from eight regions. 
Each image is a slice of the original 3D image from the SAT-DS dataset, including multiple segmentation targets marked with different colors.
The visualization demonstrates the robust capability of our decomposed model to accurately segment complex anatomical structures across diverse body regions. 
The consistent performance across different organ systems highlights the model's effectiveness in handling varied segmentation tasks.

\begin{table*}[!tbh]
\center
\caption{The 49 datasets that make up SAT-DS.
``Mulptle'' represents that the dataset contains labels from multiple body regions. 
The specific regions are marked with taskIDs, where ``1,3,5,7'' means the dataset contains labels four regions including Abdomen, Head and Neck, Pelvis, and Thorax.
}
\label{tab:dataset_details}
\vspace{0.3cm}
\setlength\tabcolsep{12pt}
\resizebox{1.0\textwidth}{!}{\begin{tabular}{lclcccccc}

\toprule
\rowcolor{lightgray} \textbf{Dataset Name} & \textbf{\#Scans} & \textbf{\#Classes} & \textbf{\#Annotations}  & \textbf{Region} \\
\midrule

\rowcolor{lightgray!50} \textbf{CT Data}           &              &        &        &               \\
AbdomenCT1K~\cite{AbdomenCT1K} & 988      & 4      &3,950    & Abdomen       \\
ACDC~\cite{ACDC}           & 300          & 4      &1,200        & Thorax        \\
AMOS22 CT~\cite{AMOS22}           & 300          & 16      & 4,765       & Abdomen        \\
Couinaud~\cite{Couinaud} & 161 & 10   &1,599        & Abdomen       \\
CrossMoDA2021~\cite{CrossMoDA2021}           & 105          & 2      & 210       & Head and Neck        \\
CT-ORG~\cite{CTORG}           & 140          & 6      & 680       & Multiple (1,3,5,6)       \\
CTPelvic1K~\cite{CTPelvic1K}           & 117          & 5      & 585       & Lower Limb        \\
FLARE22~\cite{FLARE22}     & 50           & 15     & 750        & Abdomen       \\
FUMPE~\cite{FUMPE}           & 35          & 1      & 33       & Thorax        \\
HAN Seg~\cite{HANSeg}      & 41           & 41     &1,681        & Head and Neck \\
INSTANCE~\cite{INSTANCE}           & 100          & 1      & 100       & Brain        \\
KiPA22~\cite{KiPA22}           & 70          & 4      & 280       & Abdomen        \\
KiTS23~\cite{KiTS23}           & 489          & 3      & 1226       & Abdomen        \\
LNDb~\cite{LNDb}           & 236          & 1      & 206       & Thorax        \\
LUNA16~\cite{LUNA16}       & 888          & 4      & 3,551        & Thorax        \\
MM-WHS CT~\cite{MMWHS}           & 40          & 9      & 180       & Thorax        \\
NSCLC~\cite{NSCLC}         & 85           & 2      &162        & Thorax        \\
Pancreas CT~\cite{PancreasCT}           & 80          & 1      & 80       & Abdomen        \\
Parse2022~\cite{PARSE2022}           & 100          & 1      & 100       & Thorax        \\
PDDCA~\cite{PDDCA}           & 48          & 12      & 543       & Head and Neck        \\
SEGA~\cite{SEGA}           & 56          & 1      & 56       & Abdomen       \\
SegRap2023 Task 1~\cite{SegRap2023}           & 120          & 61      & 7320       & Head and Neck        \\
SegRap2023 Task 2~\cite{SegRap2023}           & 120          & 2      & 240       & Head and Neck        \\
SegTHOR~\cite{SegTHOR}           & 40          & 4      & 160       & Thorax        \\
SLIVER07~\cite{SLIVER07}   & 20           & 1      &20        & Abdomen       \\
TotalSegmentor Cardiac~\cite{Totalsegmentator} & 1,202 & 17 &13,264       & Multiple (1,5,7)    \\
TotalSegmentor Muscles~\cite{Totalsegmentator} & 1,202 & 31 &21,510       & Multiple (3,4,5,7)    \\
TotalSegmentor Organs~\cite{Totalsegmentator} & 1,202 & 24 &20,361       & Multiple (1,3,5,7)  \\
TotalSegmentor Ribs~\cite{Totalsegmentator} & 1,202 & 39 &32,666       & Thorax    \\
TotalSegmentor Vertebrae~\cite{Totalsegmentator} & 1,202 & 29 &19,503       & Spine    \\
VerSe~\cite{VerSe} & 96 & 29 & 1,295      & Spine    \\
WORD~\cite{WORD}           & 150          & 18     &2,700        & Abdomen       \\
\rowcolor{lightgray!50} \textbf{MRI Data}          &              &        &        &               \\
AMOS22 MRI~\cite{AMOS22}           & 60          & 16      & 896       & Thorax        \\
ATLAS~\cite{ATLAS}           & 60          & 2      & 120       & Abdomen        \\
ATLASR2~\cite{ATLASR2}           & 654          & 1      & 652       & Brain        \\
Brain Atlas~\cite{Brain_Atlas} & 30       & 108    &3,240        & Brain         \\
BrainPTM~\cite{BrainPTM}   & 60           & 7      &408        & Brain         \\
CHAOS MRI~\cite{CHAOS}         & 60           & 5      &300        & Abdomen       \\
CMRxMotion~\cite{CMRxMotion}           & 138          &   4    &   536     & Thorax        \\
FeTA2022~\cite{FeTA2022}           & 80          &   7   &   560  & Brain        \\
ISLES2022~\cite{ISLES2022} & 500          & 1      &492        & Brain         \\
LAScarQS2022 Task 1~\cite{LAScarQS2022}           & 60          &   2   &  120   & Thorax        \\
LAScarQS2022 Task 2~\cite{LAScarQS2022}           & 130          &  1    &  130   & Thorax        \\
MRSpineSeg~\cite{MRSpineSeg} & 91         & 23     &1,783        & Spine         \\
MM-WHS MRI~\cite{MMWHS}           & 40          &    9  &  180  & Thorax        \\
MyoPS2020~\cite{MyoPS2020}           & 135          &   6   &  450   & Thorax        \\
PROMISE12~\cite{PROMISE12} & 50           & 1      &50        & Pelvis        \\
SKI10~\cite{SKI10}         & 99           & 4      &396        & Upper Limb    \\
WMH~\cite{WMH} & 170 & 1    &170        & Brain         \\ 
\rowcolor{lightgray!50} \textbf{Summary}           &  13,303      & 428     & 214,816 & /             \\
\bottomrule
\end{tabular}}
\end{table*}

\begin{table*}[!tbh]
\center
\caption{The configurations of 49 nnU-Nets trained on each dataset. All nnU-Nets are planned under the ``3d\_fullres'' setting. The total size of all the nnU-Nets is around 1545M.}
\label{tab:nnunet_config}
\resizebox{1.0\textwidth}{!}{\begin{tabular}{lclcccccc}

\toprule
\rowcolor{lightgray} \multicolumn{1}{l}{\textbf{Dataset}} & \multicolumn{1}{c}{\textbf{Input Size}} & \multicolumn{1}{c}{\textbf{\#Stage}} & \multicolumn{1}{c}{\textbf{\#Depth }} & \multicolumn{1}{c}{\textbf{\#Width}} & \multicolumn{1}{c}{\textbf{Model Size}} \\ 
\midrule

AbdomenCT1K~\cite{AbdomenCT1K}                     & [96 160 160]  & 6 & [2 2 2 2 2 2]   & [32 64 128 256 320 320]     & 31M \\
ACDC~\cite{ACDC}                                   & [10 256 224]  & 6 & [2 2 2 2 2 2]   & [32 64 128 256 320 320]     & 31M \\
AMOS22 CT~\cite{AMOS22}                            & [64 192 160]  & 6 & [2 2 2 2 2 2]   & [32 64 128 256 320 320]     & 31M \\
AMOS22 MRI~\cite{AMOS22}                           & [64 160 224]  & 6 & [2 2 2 2 2 2]   & [32 64 128 256 320 320]     & 31M \\
ATLASR2~\cite{ATLASR2}                             & [128 128 128] & 6 & [2 2 2 2 2 2]   & [32 64 128 256 320 320]     & 31M \\
ATLAS~\cite{ATLAS}                                 & [48 192 224]  & 6 & [2 2 2 2 2 2]   & [32 64 128 256 320 320]     & 31M \\
Brain Atlas~\cite{Brain_Atlas}                     & [112 128 112] & 5 & [2 2 2 2 2]     & [32 64 128 256 320]         & 17M \\
BrainPTM~\cite{BrainPTM}                           & [112 144 112] & 5 & [2 2 2 2 2]     & [32 64 128 256 320]         & 17M \\
CHAOS MRI~\cite{CHAOS}                             & [32 192 288]  & 6 & [2 2 2 2 2 2]   & [32 64 128 256 320 320]     & 31M \\
CMRxMotion~\cite{CMRxMotion}                       & [10 448 384]  & 7 & [2 2 2 2 2 2 2] & [32 64 128 256 320 320 320] & 45M \\
Couinaud~\cite{Couinaud}                           & [64 192 192]  & 6 & [2 2 2 2 2 2]   & [32 64 128 256 320 320]     & 31M \\
CrossMoDA2021~\cite{CrossMoDA2021}                 & [48 224 192]  & 6 & [2 2 2 2 2 2]   & [32 64 128 256 320 320]     & 31M \\
CT-ORG~\cite{CTORG}                                & [128 128 128] & 6 & [2 2 2 2 2 2]   & [32 64 128 256 320 320]     & 31M \\
CTPelvic1K~\cite{CTPelvic1K}                       & [96 160 160]  & 6 & [2 2 2 2 2 2]   & [32 64 128 256 320 320]     & 31M \\
FeTA2022~\cite{FeTA2022}                           & [96 112 96]   & 5 & [2 2 2 2 2]     & [32 64 128 256 320]         & 17M \\
FLARE22~\cite{FLARE22}                             & [40 224 192]  & 6 & [2 2 2 2 2 2]   & [32 64 128 256 320 320]     & 31M \\
FUMPE~\cite{FUMPE}                                 & [80 192 160]  & 6 & [2 2 2 2 2 2]   & [32 64 128 256 320 320]     & 31M \\
HAN Seg~\cite{HANSeg}                              & [40 224 192]  & 6 & [2 2 2 2 2 2]   & [32 64 128 256 320 320]     & 31M \\
Instance22~\cite{INSTANCE}                         & [16 320 320]  & 7 & [2 2 2 2 2 2 2] & [32 64 128 256 320 320 320] & 45M \\
ISLES2022~\cite{ISLES2022}                         & [80 96 80]    & 5 & [2 2 2 2 2]     & [32 64 128 256 320]         & 17M \\
KiPA22~\cite{KiPA22}                               & [160 128 112] & 6 & [2 2 2 2 2 2]   & [32 64 128 256 320 320]     & 31M \\
KiTS23~\cite{KiTS23}                               & [128 128 128] & 6 & [2 2 2 2 2 2]   & [32 64 128 256 320 320]     & 31M \\
LAScarQS22 Task1~\cite{LAScarQS2022}               & [24 256 256]  & 7 & [2 2 2 2 2 2 2] & [32 64 128 256 320 320 320] & 45M \\
LAScarQS22 Task2~\cite{LAScarQS2022}               & [40 256 224]  & 6 & [2 2 2 2 2 2]   & [32 64 128 256 320 320]     & 31M \\
LNDb~\cite{LNDb}                                   & [96 160 160]  & 6 & [2 2 2 2 2 2]   & [32 64 128 256 320 320]     & 31M \\
LUNA16~\cite{LUNA16}                               & [80 192 160]  & 6 & [2 2 2 2 2 2]   & [32 64 128 256 320 320]     & 31M \\
MM-WHS CT~\cite{MMWHS}                             & [80 192 160]  & 6 & [2 2 2 2 2 2]   & [32 64 128 256 320 320]     & 31M \\
MM-WHS MR~\cite{MMWHS}                             & [96 160 160]  & 6 & [2 2 2 2 2 2]   & [32 64 128 256 320 320]     & 31M \\
MRSpineSeg~\cite{MRSpineSeg}                       & [8 640 320]   & 7 & [2 2 2 2 2 2 2] & [32 64 128 256 320 320 320] & 43M \\
MyoPS2020~\cite{MyoPS2020}                         & [48 224 224]  & 6 & [2 2 2 2 2 2]   & [32 64 128 256 320 320]     & 31M \\
NSCLC~\cite{NSCLC}                                 & [48 224 192]  & 6 & [2 2 2 2 2 2]   & [32 64 128 256 320 320]     & 31M \\
Pancreas CT~\cite{PancreasCT}                       & [80 192 160]  & 6 & [2 2 2 2 2 2]   & [32 64 128 256 320 320]     & 31M \\
PARSE2022~\cite{PARSE2022}                         & [96 160 160]  & 6 & [2 2 2 2 2 2]   & [32 64 128 256 320 320]     & 31M \\
PDDCA~\cite{PDDCA}                                 & [48 192 192]  & 6 & [2 2 2 2 2 2]   & [32 64 128 256 320 320]     & 31M \\
PROMISE12~\cite{PROMISE12}                         & [20 320 256]  & 7 & [2 2 2 2 2 2 2] & [32 64 128 256 320 320 320] & 45M \\
SEGA~\cite{SEGA}                                   & [72 160 160]  & 6 & [2 2 2 2 2 2]   & [32 64 128 256 320 320]     & 31M \\
SegRap2023 Task1~\cite{SegRap2023}             & [28 256 224]  & 6 & [2 2 2 2 2 2]   & [32 64 128 256 320 320]     & 31M \\
SegRap2023 Task2~\cite{SegRap2023}             & [28 256 256]  & 7 & [2 2 2 2 2 2 2] & [32 64 128 256 320 320 320] & 45M \\
SegTHOR~\cite{SegTHOR}                             & [64 192 160]  & 6 & [2 2 2 2 2 2]   & [32 64 128 256 320 320]     & 31M \\
SKI10~\cite{SKI10}                                 & [64 192 160]  & 6 & [2 2 2 2 2 2]   & [32 64 128 256 320 320]     & 31M \\
SLIVER07~\cite{SLIVER07}                           & [80 192 160]  & 6 & [2 2 2 2 2 2]   & [32 64 128 256 320 320]     & 31M \\
TS Heart~\cite{Totalsegmentator}     & [128 128 128] & 6 & [2 2 2 2 2 2]   & [32 64 128 256 320 320]     & 31M \\
TS Muscles~\cite{Totalsegmentator}   & [128 128 128] & 6 & [2 2 2 2 2 2]   & [32 64 128 256 320 320]     & 31M \\
TS Organs~\cite{Totalsegmentator}    & [128 128 128] & 6 & [2 2 2 2 2 2]   & [32 64 128 256 320 320]     & 31M \\
TS Ribs~\cite{Totalsegmentator}      & [128 128 128] & 6 & [2 2 2 2 2 2]   & [32 64 128 256 320 320]     & 31M \\
TS Vertebrae~\cite{Totalsegmentator} & [128 128 128] & 6 & [2 2 2 2 2 2]   & [32 64 128 256 320 320]     & 31M \\
VerSe~\cite{VerSe}                                 & [160 128 112] & 6 & [2 2 2 2 2 2]   & [32 64 128 256 320 320]     & 31M \\
WMH~\cite{WMH}                                     & [48 224 192]  & 6 & [2 2 2 2 2 2]   & [32 64 128 256 320 320]     & 31M \\
WORD~\cite{WORD}                                   & [64 192 160]  & 6 & [2 2 2 2 2 2]   & [32 64 128 256 320 320]     & 31M \\
\bottomrule
\end{tabular}}
\end{table*}

\bibliographystyle{ieeetr}
\bibliography{supple}
